\PassOptionsToPackage{dvipsnames}{xcolor}
\documentclass{article}

\usepackage[preprint]{neurips_2026}
\usepackage[utf8]{inputenc}
\usepackage[T1]{fontenc}
\usepackage{hyperref}
\usepackage{url}
\usepackage{booktabs}
\usepackage{amsfonts}
\usepackage{nicefrac}
\usepackage{microtype}
\usepackage{xcolor}
\usepackage{amsmath}
\usepackage{grffile}
\usepackage{caption}
\usepackage{tabularx}
\usepackage{seqsplit}
\usepackage{array}
\usepackage{makecell}
\usepackage{algorithm}
\usepackage[noend]{algpseudocode}
\usepackage[normalem]{ulem}
\usepackage{subcaption}
\usepackage{cleveref}
\usepackage{glossaries}

\usepackage[dvipsnames]{xcolor}

\usepackage{algorithm}
\usepackage[noend]{algpseudocode}
\usepackage{amsmath,amsthm,amsfonts}
\usepackage[normalem]{ulem}
\usepackage{subcaption}

\newcommand{\R}{\mathbb{R}}

\usepackage{cleveref}

\crefname{section}{Section}{Sections}
\crefname{subsection}{Section}{Sections}
\crefname{subsubsection}{Section}{Sections}
\crefname{appendix}{Appendix}{Appendices}
\crefname{table}{Table}{Tables}

\newtheorem{theorem}{Theorem}

\newtheorem{corollary}{Corollary}

\theoremstyle{definition}
\newtheorem{definition}{Definition}

\crefname{theorem}{Theorem}{Theorems}
\crefname{lemma}{Lemma}{Lemmas}
\crefname{claim}{Claim}{Claims}
\crefname{conjecture}{Conjecture}{Conjectures}
\crefname{proposition}{Proposition}{Propositions}
\crefname{definition}{Definition}{Definitions}
\crefname{remark}{Remark}{Remarks}
\crefname{equation}{Eqn.}{Eqns.}

\usepackage{glossaries}

\newacronym{frameworkname}{PEM}{Personalized Estimation Modeling}

\title{Why Global LLM Leaderboards Are Misleading: Small Portfolios for Heterogeneous Supervised ML}

\author{%
  Jai Moondra\thanks{Corresponding authors: Jai Moondra (\texttt{jaimoondra@cmu.edu}) and Swati Gupta (\texttt{swatig@mit.edu})} \thanks{Equal contribution}  \\
  Carnegie Mellon University\\
  \texttt{jaimoondra@cmu.edu} \\
  \And
  Ayela Chughtai\footnotemark[\value{footnote}] \\
  MIT Sloan School of Management \\
  \texttt{ayela@mit.edu} \\
  \And
  Bhargavi Lanka\thanks{A part of this work was done when the author was a student at MIT Sloan School of Management.} \\
  MIT Sloan School of Management \\
  \texttt{bharg@alum.mit.edu} \\
  \And
  Swati Gupta$^*$ \\
  MIT Sloan School of Management \\
  \texttt{swatig@mit.edu} \\
}

\begin{document}

\maketitle

\begin{abstract}
Ranking large language models (LLMs) via pairwise human feedback underpins current leaderboards for open-ended tasks, like creative writing and problem solving. We analyze $\sim$ 89K comparisons across 116 languages across 52 LLMs, from the Arena platform, and show that the best fit global Bradley-Terry (BT) ranking is largely misleading. Nearly two-thirds of the decisive votes cancel each other out, and even the top 50 models according to the global BT ranking are statistically indistinguishable (pairwise win probabilities are at most 0.53 within the top 50, i.e., near random outcomes). We trace this failure to strong, structured heterogeneity of opinions across language, task, and time. Moreover, we find an important characteristic - \emph{language} plays an important role in this heterogeneity. Grouping by language (and language families) increases the agreement of votes massively, resulting in two orders of magnitudes higher spread in the ELO scores (i.e., very consistent rankings). What appears as global noise is in fact a mixture of coherent but conflicting subpopulations. 
    
To address such a heterogeneity in supervised machine learning, broadly, we introduce the framework of \emph{$(\lambda, \nu)$-portfolios}, which are small collections of models that achieve a prediction error at most $\lambda$, ``covering'' at least a $\nu$ fraction of users. We formulate this problem as a variant of the set cover problem, and provide guarantees by invoking the VC dimension of the underlying set system. On the Arena dataset, our algorithms recover just $5$ distinct BT rankings that cover over 96\% of votes at a modest $\lambda$, compared to the 21\% coverage achieved by global ranking at the same threshold. We also provide a portfolio of 6 LLMs that cover twice as many votes as compared to choosing the top-$6$ LLMs from a global ranking. We further illustrate small portfolios for a classification problem on the COMPAS dataset using an ensemble of fairness-regularized classification models, and show that these portfolios can be used to detect blind spots or errors in the data, which might be of independent interest to policymakers.
\end{abstract}

\section{Introduction}\label{sec: introduction}

Large language models (LLMs) are increasingly deployed at scale to serve a global and diverse user base. Modern approaches evaluate them on specific tasks like creative writing, complexity, creativity, domain knowledge, problem solving, specificity, technical accuracy, code, math, and real-world use cases. As response quality from these models varies with prompt formatting and dataset contamination \citep{sclar2023quantifying, voronov2024mind, alzahrani2024benchmarks}, evaluation of LLMs through \emph{human feedback} is often preferred. Moreover, recent studies highlight that human feedback can capture diversity more effectively than standard benchmarks \citep{kirk2024prism}. Open evaluation platforms like \href{arena.ai}{Arena} \citep{pmlr-v235-chiang24b, arena2026} (formerly, LMArena) rank large language models based on human feedback (in terms of preference amongst two displayed responses), and have become a transparent gold standard for assessing LLMs \citep{wsj_ucb_ai_obsession_2024}. However -- as we show -- due to heterogeneity of the votes in the dataset, the global ranking constructed by such platforms is {\it not representative of the user votes and opinions across the world}. On the recent publicly available data from HuggingFace\footnote{\url{https://huggingface.co/datasets/lmarena-ai/arena-human-preference-140k}} on 52 LLMs\footnote{See a list of names in Table \ref{tab:llm-mapping} in Appendix \ref{sec:llm-mapping}.} across 116 languages, we show that 72.28\% of decisive non-tied votes (across $\sim 89k$ in the entire data) are actually `canceled' \footnote{That is, \% of individual decisive votes that are offset by an opposing decisive vote for the same model pair, where one vote favors LLM 1 over LLM 2 and another favors LLM 2 over LLM 1.} by each other.

Moreover, since users only provide comparative feedback among a pair of LLM responses to their prompt, platforms like Arena convert votes to a global preference using the standard Bradley-Terry (BT) ranking model (\citep{bradley1952rank}, also widely used for routing queries to the most appropriate LLM or inference-time scaling \cite{damanilearning}). This model estimates an \emph{Elo score} $\theta_\ell \in \R$ for each LLM $\ell$ (see Section \ref{sec: preliminaries}), with higher scores indicating a greater probability that the user base prefers $\ell$ to other LLMs. Despite their usefulness in extrapolating pairwise feedback to a ranking, these models can end up erasing information about conflicting preferences or be dominated by the larger group's preference. This leads to a peculiar \emph{hedging behavior}, wherein the global model may have higher error across groups with different trends. In particular, for Arena's case, we find that the global ranking is able to predict the winner of a vote with $\ge 70\%$ probability for only $10.3 \%$ of all votes.\footnote{Note that these are not unseen/test votes -- these were used to train the global BT ranking (see Section \ref{sec: arena}). While some of this `hedging' is by design -- Arena's algorithm \citep{pmlr-v235-chiang24b} is more likely to match similarly ranked models more often to be able to separate them -- it does not explain varied preferences across language groups.} %Thus, one can obtain an ordering of different LLMs using only pairwise comparisons, and such \emph{global} rankings are useful in assessing real-world performance of models. More broadly, there is a vast literature on BT rankings for summarizing user preferences across \jai{(Cite)}.  
  %27.37\% of the votes are ties across models, 

\begin{figure}[t]
    \centering
    \includegraphics[width=\textwidth]{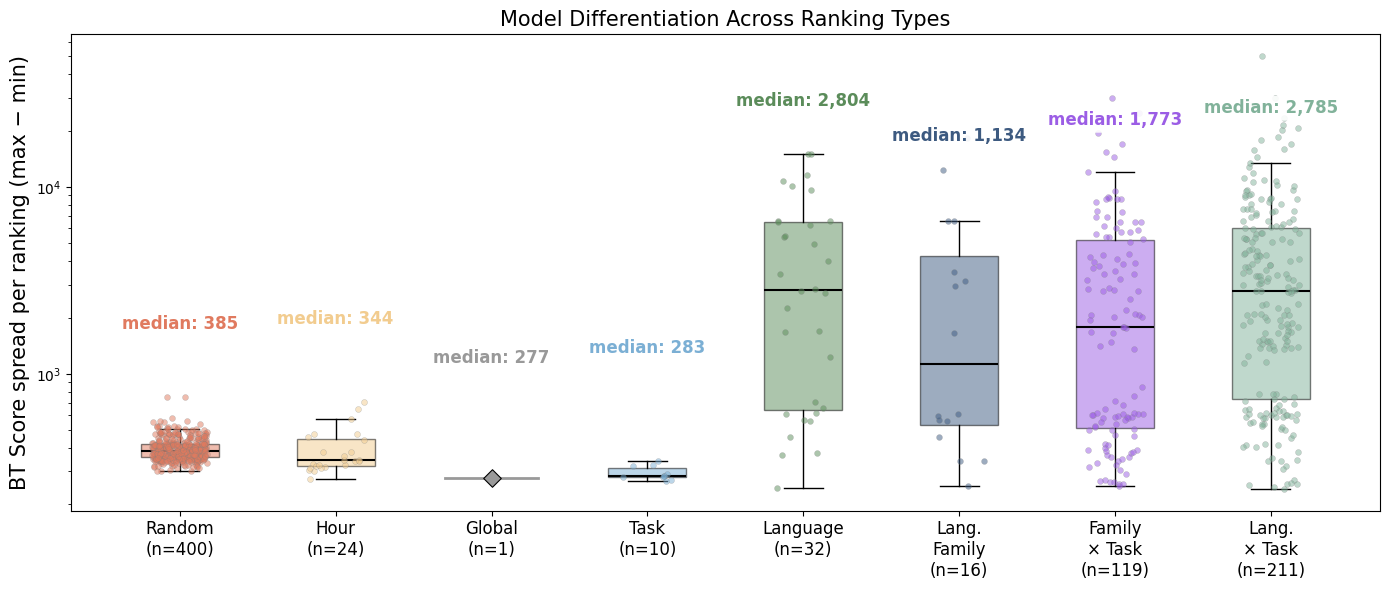}
    \caption{Model differentiation across ranking types. Each point is a BT ranking fitted on a group of votes ($< 50$ votes excluded). The y-axis reports Elo-scaled score spread (max \-- min; see Eqn.~\ref{eqn: elo-score} and Table~\ref{tab: elo-score-to-win-probability}), where higher spread indicates more aligned preferences and more predictive rankings. Language, language-family, and their task interactions induce substantially larger spreads than random, hourly, task-only, or global rankings.}\label{fig:full_score_spread}
    \vspace{-0.5cm}
\end{figure}

We further note that even the highest-ranked models in the global BT ranking exhibit substantial variation across language family BT rankings. Among the global top five models, Gemini-2.5P, Gemini-2.5P3, Grok-4, o3, and Gemini-2.5-P5 differ from their global ranks by 5 to 14 positions on average across rankings restricted to votes from specific language-families; in fact Grok-4 even falls to rank 52 in the Afro-Asiatic family ranking. On the Arena website, e.g., as of May 4th, there is only a 50 point Elo score gap (see Section \ref{sec: preliminaries}) between top 50 models, implying a probability of winning of 0.53, of the top 1 model winning against the rank 50 model. Recent parallel work that also shows that these rankings are unreliable due to variance under vote dropout of <5\% \cite{huang2025dropping}.

\begin{figure}[t]
    \centering
    \includegraphics[width=\textwidth]{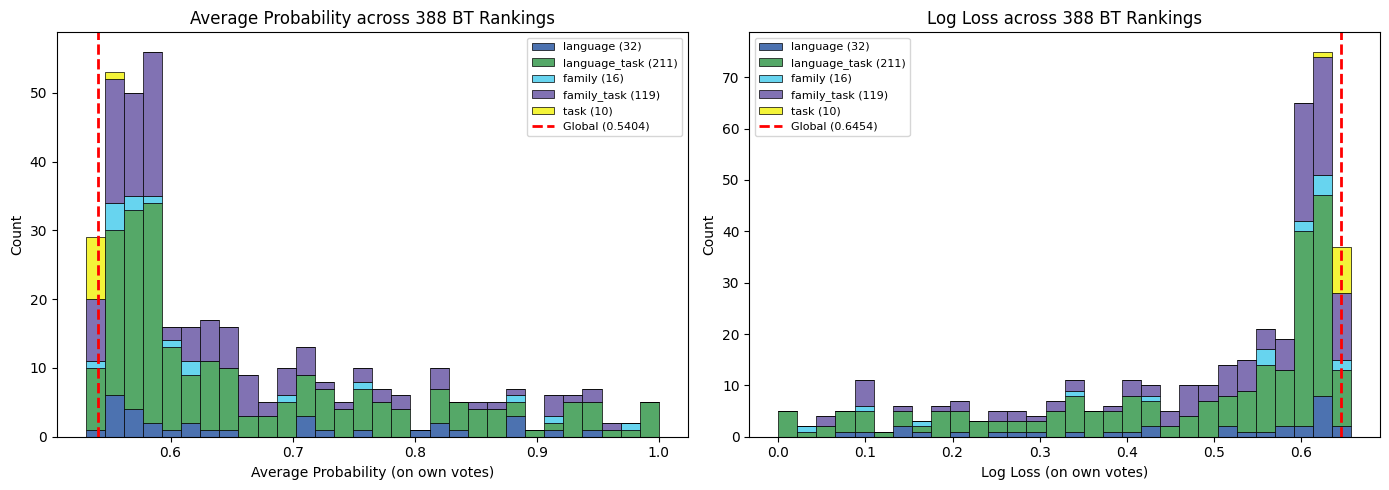}
    \caption{ 
    In-group predictive performance of 388 subpopulation BT rankings vs.\ the global ranking. Left: average probability assigned to the observed winner. Right: average log loss. Colors indicate subpopulation type. Language- and family-based rankings consistently assign sharper winner probabilities and lower log loss than the global ranking, while task-specific rankings remain non-representative, similar to the global baseline.
    }
    \label{fig:subpop-bt-performance}
    \vspace{-0.5cm}
\end{figure}

In this work, we dig deeper into the characteristics of heterogeneity in user preferences across the votes collected by Arena. In particular, we trace this failure to strong, structured heterogeneity of opinions across language, task, and time. However, we find an important characteristic - {\it language plays an important role} in this heterogeneity. Grouping by language (and language families) increases the agreement of votes massively, resulting in much lower log-loss per group and a much higher probability of sampling the votes within each group (see Figure \ref{fig:subpop-bt-performance}). What appears as global noise is in fact a {\it mixture of coherent but conflicting subpopulations}.

This phenomenon of `best-fit' models hedging between heterogeneous groups and losing all statistical or predictive significance is observed more generally in all supervised learning problems. For example, Simpson's paradox \cite{simpson1951interpretation} arises due to the reversal of certain trends in different groups of data when these groups are combined. We provide a way forward by proposing an alternate: to find a set of supervised ML models that together achieve a smaller loss on a diverse population. For the Arena case, we find a small set of rankings that critically use `more homogeneous clusters' within the votes based on the prompt task, language, and language-family, thus solving the key challenge faced in this area. Our general framework addresses the following question: 

\begin{center}
\emph{
    Given a large ensemble of ML models (or rankings) in a supervised learning setting, can we select a small subset or \emph{portfolio} of models such that each user (or vote) is reasonably ``covered'' by at least one model in the portfolio?
}
\end{center}

{\bf Outline and Summary of Key Contributions:} In this work, we formalize this trade-off for supervised learning, and present an algorithmic framework for choosing small collections or \emph{portfolios} \citep{gupta_which_2023, kim_navigating_2025} of ML models that ensure that most (or all) users are satisfactorily `covered' by some ML model in the portfolio (Section \ref{sec: framework}). Specifically, we show that the corresponding problem is an instance of the well-known ``partial set cover problem'' \citep{karp1975, kearns1990computational, slavik1997improved}. Next, in Section \ref{sec:algorithms}, we present algorithms with provable guarantees on the (approximately) smallest number of models that cover enough users. 

By exploiting the algorithmic connection between set cover and the VC dimension of the underlying set system \cite{bronnimann1994almost}, we obtain novel guarantees on portfolio sizes for regression problems, and in particular for linear regression (Thm \ref{thm: faster-set-cover}), and discuss the extension to ranking models (Section \ref{sec:arena-prediction}). 

Finally, we discuss our computational case studies on Arena and COMPAS in Section \ref{sec: arena}.  For the Arena dataset (Section \ref{sec:arenacase}), we provide evidence of heterogeneity in user preferences, and homogeneity within language-based subgroups. 

We demonstrate that our framework can generate small portfolios (with only between $3$ and $10$ models each) for Arena that outperform any global model in capturing diverse users and reducing performance gaps across families. 
Similar results hold for LLMs: we show most voters are reasonably well-served by a small collection of $4$ to $8$ LLMs (see Figure \ref{fig:coverage_comp_llms}). Simply collecting the top few LLMs in the global ranking does not achieve the same effect. We further show a proof of concept for other heterogeneous supervise ML settings, by constructing portfolios to predict recidivism of criminal defendants within 2 years using the COMPAS data \cite{angwin2022machine} (Section \ref{sec:compascase}), and discuss the characteristics of the defendants that are harder to cover with small portfolios (e.g., low priors, young), highlighting the need to develop better models. 

\subsection{Related work}\label{sec: related-work}
{\it Ranking Models and Preference Aggregation.}
A large literature studies aggregating pairwise preferences into global rankings (e.g., Bradley-Terry) \cite{bradley1952rank}, but recent work highlights their limitations under heterogeneity. Group-level inconsistencies, such as those observed in hiring and subgroup fairness \cite{cachel2025}, \cite{kearns2018}, show that aggregate rankings can mask conflicting preferences. In the context of LLM leaderboards, additional concerns arise: rankings are vulnerable to manipulation via injected votes \cite{min2025improving}, strategic voting \cite{huang2025exploring}, and gaming of LLM-based evaluators \cite{zheng2025, raina2024}, as well as issues like data leakage \cite{singhleaderboard}. Complementing this line of work, we show that even without adversaries, global rankings can be inherently misleading due to structured heterogeneity, and are better viewed as a portfolio of homogeneous models so that opposing preferences are preserved.

{\it Individual Fairness.} There is a long line of work advocating {\it individual fairness}, i.e., that similar individuals should be treating similarly \cite{dwork2012, dwork2020individual, dwork2018individual, anderson2025, gupta2021individual}. There has been a lot of work on defining the metric under which similarity is defined (e.g., \cite{ilvento2020, waller2025beyond}), auditing in absence of such a metric \cite{bechavod2020metric}, and in even averaged versions of individual fairness that ask for similar classification accuracy across multiple tasks for each individual \cite{sharifi2019average}. In this work, we take the view that all individuals (i.e., the metric is zero everywhere) must enjoy an upper bound on the amount of error they receive from the model predictions, which is essential in assessing risk compliance of a tool deployed in practice. In the case of ranking LLMs, this amounts to a minimum representation provided by chosen large language models in the ranking setting for each vote. Our work therefore studies the choice of models from an ensemble that operationalizes individual fairness in terms of bounded error. This provides a way to address the tension between the two desiderata of fairness and accuracy, by asking for a bounded measurement error on each individual \cite{kinney2025}. 

{\it Ensemble Methods.} Literature on ensemble methods focuses on boosting and bagging methods for classification and regression models (e.g., \cite{dietterich2000ensemble} and ref. therein). Our work however focuses on selected a smaller subset of models from any given ensemble so that the maximum error from the {\it best} of the chosen models is bounded for each individual. Therefore, our work complements the sparsification literature in ML, where the typical motivation is to improve explainability and interpretability, e.g., \cite{liu2023fire}, where the authors use minimization of aggregate least squares loss, rather than bounding the loss on each datapoint.

\section{Preliminaries}\label{sec: preliminaries}
\textbf{Setup}.Throughout the paper, we assume a supervised learning set up, and denote by $\mathcal{X} = (x_i)$ the set of $n := |\mathcal{X}|$ data points, and by $\mathcal{Y} = (y_i)$ the corresponding set of real-valued labels. For classification problems, the label $y_i \in \{0, 1\}$ for each $i$. A model $h: \mathcal{X} \to \R$ takes each data point $x_i$ to a prediction $h(x_i)$ for its label $y_i$. For classification models, $h(x_i) \in [0, 1]$ can be interpreted as the model's estimated probability of the label $y_i$ being $1$.

\textbf{Arena.} The Arena platform collects human preferences through pairwise comparisons between LLMs. Each time a user or a \emph{voter} enters a prompt, the user is shown responses from two LLMs and can select a preferred output. Let $\mathcal{L}=\{1, \dots, L\}$ denote the set of LLMs. Each vote involves two distinct LLMs $a, b \in \mathcal{L}$ and produces one of four outcomes: $a$ wins, $b$ wins, tie, and `both bad'. We exclude votes labeled as `both bad'. We treat a tie as half a win for both LLMs. Hereafter, we assume $n$ total votes, and refer to the $i$th vote as $(a_i \succ b_i)$ where $a_i$ wins against $b_i$, for $i \in [n]$.

\textbf{Ranking model.} The Bradley-Terry (BT) ranking model \citep{bradley1952rank} associates with each LLM $a \in \mathcal{L}$ a score $\theta_\ell \in \R$, with the probability that $a$ wins against $b$ in a vote computed as
\[
    \Pr(a \succ b) = \tfrac{\exp(\theta_a)}{\exp(\theta_a)+\exp(\theta_b)} = \sigma(\theta_b-\theta_a),
\]
where $\sigma(x) = (1 + 10^{-x})^{-1}$ denotes the standard logistic function, therefore larger gap in the scores corresponds to a sharper winning probability. Some model pairs appear more frequently than others in the data. To avoid overweighting commonly sampled pairs, we use inverse-probability weighted (IPW) win counts, as prescribed in \cite{pmlr-v235-chiang24b}. Let $\hat{P}(\{a, b\})$ denote the fraction of votes comparing $a$ and $b$ in the dataset. We define the weight $w_{ab} := \sum_{\substack{\text{votes where } a \\ \text{beats } b}} \frac{1}{\hat{P}(\{a, b\})}$.

The resulting IPW log-likelihood is $\ell(\theta)
:= \sum_{a, b \in \mathcal{L}}
\Big(
w_{ab}\log \sigma(\theta_a - \theta_b)
+ w_{ba}\log \sigma(\theta_b - \theta_a)
\Big)$.

Scores $\theta_a, a \in \mathcal{L}$ are obtained by optimizing the log-likelihood.
Since winner predictions depend only on score differences, we can center the estimated scores to have zero mean.

For interpretability, we report Elo-scaled scores 
\begin{equation}\label{eqn: elo-score}
    \text{Rating}_m = 400 \cdot \theta_m + 1000
\end{equation}
that correspond to a base-10 logistic scale commonly used in Elo rating systems, where a 400-point difference implies approximately 10:1 odds of preference \citep{elo1967proposed}. 
%\textbf{Metadata.} 
Each vote in the Arena dataset contains additional data about the vote besides the names of the winning and losing LLMs. In this work, we will also use the (a) timestamp, (b) prompt/response language, and (c) \emph{task} category of the prompt (as defined by Arena). In particular, \emph{the identity of the voter is not provided}, and a single person may have cast multiple votes. For simplicity, we treat each vote as a `user'.

\section{Framework and algorithms}\label{sec: framework}

We next propose small `portfolios' of models that guarantee a reasonable `coverage' across all data points or users. We first describe it generally for any supervised learning setup, and then specifically for Arena in Section \ref{sec: arena}, where users correspond to votes.

Consider $n$ \emph{data points} $\mathcal{X} = \{x_1, \ldots, x_n\}$ with associated labels $y_1, \ldots, y_n \in \mathbb{R}$. Given a \emph{margin} $\lambda \ge 0$, we say that a model $h: \mathcal{X} \to \R$ \emph{covers} $x_i \in \mathcal{X}$ if $|h(x_i) - y_i| \le \lambda$. We next define $(\lambda, \nu)$-portfolio, which is a collection of models such that at least $\nu$ fraction of data points have at least one model in this collection that $\lambda$-covers it:

\begin{definition}[$(\lambda, \nu)$-portfolio]
    Consider a \emph{margin} $\lambda \ge 0$, a \emph{coverage fraction} $\nu \in (0, 1]$, and set $\mathcal{X}$ of $n = |\mathcal{X}|$ data points. We say that a finite subset $\mathcal{P} = \{h_1, \ldots, h_k\} \subseteq \mathcal{H}$ of models is a $(\lambda, \nu)$-\emph{portfolio} if there is a subset $\mathcal{S}_\nu \subseteq \mathcal{X}$ of $|\mathcal{S}_\nu| \ge \nu n$ data points such that
    \begin{equation}\label{eqn: portfolio-def}
        \text{for all} \ x_i \in \mathcal{S}_\nu, \ \text{there exists} \ j \in [k]: \quad |h(x_i) - y_i| \le \lambda.
    \end{equation}
\end{definition}

Lower margin $\lambda$ indicates better prediction quality. Higher $\nu$ indicates higher coverage, or fewer outliers. Ideally, we seek models with both good predictions ($\lambda \sim 0$), and high coverage (large $\nu \sim 1$). This isn't always possible when portfolio size $k = 1$, e.g., if $\nu = 1$, the above definition requires a single model with maximum error $\le \lambda$ across all data points.

At the other extreme, portfolios of size $k \le n$ exist\footnote{We assume `reasonable' classes $\mathcal{H}$ of models: each data point $x$ and label $y$, there is always some model $h \in \mathcal{H}$ with $h(x) = y$.} trivially: for each data point $i \in [n]$, choose any hypothesis $h_i \in \mathcal{H}$ that satisfies $h_i(x_i) = y_i$. Therefore, there is a three-way trade-off between margin/quality $\lambda$, coverage fraction $\nu$, and portfolio size ($k$).

\emph{We are interested in finding portfolios where size $k > 1$ but is still very small (between $2$ and $10$). In particular, does allowing a few extra models dramatically increase quality, or cover many more people?} That is, are there regimes where $k$ and $\lambda$ are small, and $\nu$ is close to $1$? This is crucial in many applications: large $\nu$ ensures inclusivity, small $k$ ensures cost-feasibility (only a few models can be trained and maintained), and small $\lambda$ ensures meaningful predictions.

\subsection{Algorithms to find minimum-size portfolio}\label{sec:algorithms}

Next, we discuss algorithms for the problem of finding the smallest $(\lambda, \nu)$-portfolio, i.e., minimizing $k$ for given $\lambda$ and $\nu$. As we have indicated, this is a \emph{covering} problem where we seek the smallest subset $\mathcal{P} \subseteq \mathcal{H}$ so that at least $\nu n$ data points are $\lambda$-covered by some model in $\mathcal{P}$. This is an instance of the \emph{Partial Set Cover} problem \citep{kearns1990computational} and we will use corresponding algorithms and results for our setting, but omit the (straightforward) formal reduction.

\textbf{Greedy algorithm} (Folklore). Given $\lambda$ and $\nu$, consider the following natural \emph{greedy} algorithm:
\begin{itemize}
    \item Initialize portfolio $\mathcal{P} \gets \emptyset$.
    \item While fewer than $\nu n$ data points have been $\lambda$-covered by some model in the portfolio, find model $h_{\mathrm{new}} \in \mathcal{H}$ that covers the most number of uncovered data points, and add $h_{\mathrm{new}}$ to $\mathcal{P}$.
\end{itemize}

We can use the results of \cite{slavik1997improved} in our setting, to derive directly that the greedy algorithm returns a $(\lambda, \nu)$-portfolio of models with size at most $O(\log \nu n) \cdot k^*$, where $k^*$ is the minimum size of a $(\lambda, \nu)$-portfolio:

\begin{theorem}[\cite{slavik1997improved}]\label{thm: partial-set-cover-log-approximation}
    The greedy algorithm is $O(\log \nu n)$-approximation for Partial Set Cover.
\end{theorem}

\textbf{Improved guarantees using VC dimension.} While the greedy algorithm returns a $(\lambda, \nu)$-portfolio with size within factor $O(\log \nu n)$ of the optimal, it does not exploit the geometry of the model class $\mathcal{H}$. \cite{bronnimann1994almost} established a fundamental connection between Set Cover and learning theory, and gave an efficient algorithm for Set Cover that returns a portfolio of size at most $O(d \log dk^*) \cdot  k^*$, where $d$ is the VC dimension of the class $\mathcal{H}$ of classifier models, or equivalently, $d$ is the VC dimension of the underlying set system. To extend this to regression models, let us first reduce the regression problem to a classification problem:
\begin{definition}[Margin classification]
    Given a regressor $h: \mathcal{X} \to \R$ model for data points $\mathcal{X} = (x_i)$ with corresponding labels $\mathcal{Y} = (y_i)$, and a margin $\lambda \ge 0$, define the corresponding classifier $\hat{h}: \mathcal{X} \to \R$ as the indicator for whether $h$ correctly predicts $y_i$ within margin $\lambda$. That is,
    \[
     \hat{h}(x_i) = \begin{cases}
         1 & \mathrm{if} \ |h(x_i) - y_i| \le \lambda, \\
         0 & \mathrm{otherwise.} 
     \end{cases}
    \]
    For a class $\mathcal{H}$ of regression models for dataset $\mathcal{X}, \mathcal{Y}$, define the class $\hat{\mathcal{H}} := \{\hat{h}: h \in \mathcal{H}\}$ of corresponding classifier models.
\end{definition}

The problem of determining the minimum $(\lambda, \nu = 1)$ portfolio is then the same as the {\it set cover} problem for set system with elements $\mathcal{X} = (x_i)$ and sets $S_h := \{x_i \in \mathcal{X}: \hat{h}(x_i) = 1\}$ for all $h \in \mathcal{H}$. Since the VC dimension of this set system is the same as the VC dimension of the corresponding hypothesis class $\hat{\mathcal{H}}$, we get that the following result follows from \cite{bronnimann1994almost}:

\begin{theorem}\label{thm: faster-set-cover}
    There is an efficient algorithm that given a class $\mathcal{H}$ of regression models, margin $\lambda > 0$, and data points $(x_i, y_i), i \in [n]$, returns a $(\lambda, 1)$-portfolio of size at most $O(d k^* \cdot \ln (d k^*))$, where $k^*$ is the size of the smallest $(\lambda, 1)$-portfolio, and $d$ is the VC dimension of the corresponding class $\hat{\mathcal{H}}$ of margin-classifiers.
\end{theorem}

As an immediate corollary, we get the following:

\begin{corollary}
    For $d$-dimensional data points $\mathcal{X} \subseteq \R^d$, and for the class $\mathcal{H}$ of all $d$-dimensional linear regressors (hyperplanes), a $(\lambda, 1)$-portfolio of size at most $O(dk^* \cdot \ln (d k^*))$ can be found efficiently.
\end{corollary}

\textbf{Integer program.} When the model class $\mathcal{H}$ is finite,\footnote{Most model classes used in practice can be assumed to be finite using a fine-enough discretization of the parameter space.} we can formulate an integer program (IP) for this problem. The optimal solution to this integer program gives the smallest $(\lambda, \nu)$-portfolio, if such a portfolio exists. For each hypothesis $h_j \in \mathcal{H}$, binary variable $t_j \in \{0, 1\}$ indicates whether $h_j$ is included in the portfolio. We seek to minimize the total number of models included in the portfolio, i.e., $\sum_j t_j$.  For each data point $x_i \in \mathcal{X}$, binary variable $u_i \in \{0, 1\}$ indicates whether $x_i$ is covered by some model in the portfolio. Then, we must have that $\sum_{j: |h_j(x_i) - y_i| \le \lambda} t_j \ge u_i$. This gives the following IP:
\begin{align}
\vspace{-0.5cm}
    \min &\sum_{j} t_j & \text{s.t.} \label{ip: obj}\tag{MIP} \\
    \sum_{i} u_i &\ge \nu n, \label{ip: cover-enough-points} \\
    u_i &\le {\sum}_{j: |h_j(x_i) - y_i| \le \lambda} t_j & \forall \ i, \label{ip: model-covers-point-if} \\
    u_i, t_j &\in \{0, 1\} & \forall \ i, \forall \ j. \notag
    % t_j &\in \{0, 1\} & \forall \ j. \notag
\end{align}
% The objective (\ref{ip: obj}) seeks to minimize the portfolio size. 
Constraint (\ref{ip: cover-enough-points}) enforces that $\ge \nu n$ points are covered by the portfolio.

While solving this IP requires exponential time in the worst-case, it is often practical when the number of data points $n$ and the number of models $|\mathcal{H}|$ is reasonably small. For larger problems, we can relax the integer constrains and allow variables $u_i, t_j$ to lie in $[0, 1]$, obtaining a linear program and using a reasonable rounding scheme; see Appendix \ref{sec: mip-implementation-details} for details. Despite having much higher runtime than the greedy algorithm, the IP formulation often produces smaller portfolios in our experiments on the Arena dataset (Section \ref{sec: arena}). 
% Constraint (\ref{ip: model-covers-point-if}) ensures that if a point is covered, then there is some model $M_j$ in the portfolio such that $s(M_j, p_i) \le \lambda$.

\subsection{Arena as a prediction problem}\label{sec:arena-prediction}
Next, we describe this setup for Arena specifically, where data points correspond to votes. For different votes in the dataset, we wish to (1) find `good' BT ranking(s) that are good predictors of the winner of the votes, and (2) find `good' LLM(s) that are preferred by different voters.

We view BT rankings as (probabilistic) classifiers for the votes, with the `error' $\mathrm{err}_i(h)$ of a BT ranking $h$ for a vote $x_i = (a_i \succ b_i)$ defined as the probability that BT ranking assigns to the loser ($b_i$) winning the vote. Then, $\mathrm{err}_i(h) \in [0, 1]$, and smaller values correspond to a better prediction.

As before, we say that the BT ranking $\lambda$-covers this vote if $\mathrm{err}_i(h) \le \lambda$. A $(\lambda, \nu)$-portfolio is then a (finite) subset of the BT rankings such that for at least a fraction $\nu$ of all votes, there is some BT ranking in the portfolio that correctly predicts the winner of the vote within error margin $\lambda \in [0, 1]$. Informally, this indicates that the preferences of the voter-base is always captured by one of these BT rankings within a given margin, apart from $(1 - \nu) \times (\# \mathrm{total \ votes)}$ outliers. E.g., a $(0.1, 0.99)$-portfolio of $5$ rankings indicates that the preferences of $99\%$ of the users is captured by $1$ of these $5$ rankings, up to a $90\%$ confidence.

Measuring the probability that an LLM is `preferred' by a voter that cast a certain vote is trickier, since each vote consists of pairwise information among only $2$ LLMs. We extrapolate this voter's preferences to other LLMs using BT rankings, as described next. First, we compute a very large ensemble of BT rankings, each corresponding to a different subgroup of the votes based on language and task subfamilies. Then, we extrapolate the voter's preferences to other based on BT rankings based on the assumption that if a BT ranking predicts a vote `accurately', then the voter prefers higher-ranked LLMs in this BT ranking to the lower-ranked ones. We formalize this in Appendix \ref{app: bt-rankings-to-llm-scores}, and discuss this further in Section \ref{sec: arena}.

\section{Computational case studies: Arena and COMPAS}\label{sec: arena}

We next present computational results on the Arena and COMPAS datasets. All computations were performed on an Apple MacBook computer with Apple M5 (base) processor, and 32 GB RAM.

\begin{figure}[t]
    \centering
    \begin{minipage}{0.8\textwidth}
    \includegraphics[width=\linewidth]{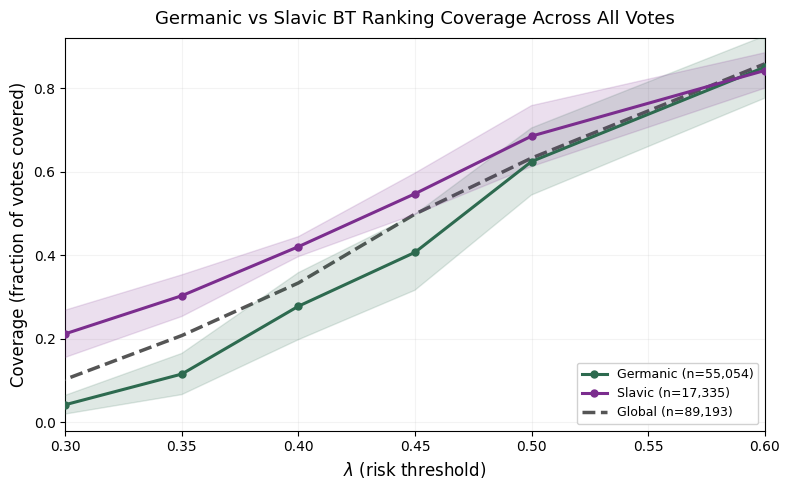}
    \end{minipage}
    \begin{minipage}{\textwidth}
        \caption{Language-family BT rankings capture distinct preference structure beyond the global ranking. Coverage is evaluated over all decisive Arena votes as we vary the $\lambda$-risk threshold. The shaded regions show approximate 95\% confidence bands around the family mean, computed across languages within each family (showing correlations within language family). Differences between the Germanic and Slavic curves indicate that the global ranking covers some language-family preferences better than others, with enough structure within language groups.}
        \label{fig:germanicslavic}
    \end{minipage}\vspace{-0.5cm}
\end{figure}

\subsection{Arena dataset with pairwise LLM comparisons}\label{sec:arenacase}

We describe our experimental setup for Arena. First, we fit a single `global' BT model on all {89,193} decisive votes (winners assigned weight 1), and {19,375} ties (both models assigned weight 0.5).  Next, we stratify the Arena human-preference data by five metadata partitions:
\emph{Language}, \emph{Language Family}, \emph{Task}, \emph{Language Family $\times$ Task} and \emph{Language $\times$ Task}. The 116 languages are mapped to coarse families (Germanic, Romance, Slavic, Indo-Iranian, Sino-Tibetan, Turkic, Uralic, Afro-Asiatic, Niger-Congo, Austronesian, Dravidian, etc; see Section \ref{sec:language-family}). Each prompt is also associated with one or more of 10 task categories (creative writing, complexity, creativity, domain knowledge, problem solving, real world, specificity, technical accuracy, code, and math). Each vote can belong to multiple subpopulations. To avoid sampling errors, we exclude any subpopulation with fewer than 50 votes. For each eligible subpopulation, we fit a standard BT model and each resulting ranking is evaluated on its \emph{own} user slice. For instance, for a language BT ranking on German, coverage is computed only on the subsets of votes that use queries written in German. We recompute $\lambda$-coverage for each new BT ranking.

\textbf{Results.} The global BT ranking leaves a large fraction of individual votes poorly explained, and predicts the winner of only $\sim 11\%$ of all votes correctly with a $> 70\%$ confidence.

For subpopulation-specific BT rankings, many subpopulations achieve \emph{much higher} coverage than the global baseline when evaluated on their own data. Further, language family-level BT rankings are often also useful for languages with fewer than $50$ votes for which we do not compute a dedicated BT ranking. Several language–task strata (e.g., (English, code) vs. (German, creative)) show different top models than the global ranking. This indicates heterogeneity across the whole population but relative homogeneity within subpopulations (see Figure \ref{fig:heterogeneity} for rankings by some languages).
% \sg{Where can readers view these BT rankigns?}

- \emph{Portfolio of BT rankings.} From our initial ensemble of \textbf{389} BT rankings, we seek a \emph{small} portfolio that maximizes vote-level $\lambda$-coverage for margin $\lambda \in \{0.05, 0.10, \ldots, 0.45, 0.50\} \cup \{0.6, 0.7, 0.8, 0.9\}$. As before, a BT ranking $\lambda$-covers a vote $(a \succ b)$ if it predicts that $a$ wins a vote against $b$ with probability $\ge 1 - \lambda$ (equivalently, from Eqn. \ref{eqn: elo-score}, the Elo scores for this ranking satisfy $\mathrm{score}_a - \mathrm{score}_b \ge \ln \frac{1 - \lambda}{\lambda}$). Given some coverage fraction $\nu \in (0, 1]$, this sets up the portfolio optimization problem.

\begin{figure}[t]
    \centering
    \begin{subfigure}[t]{\textwidth}
        \centering
        \includegraphics[width=\textwidth]{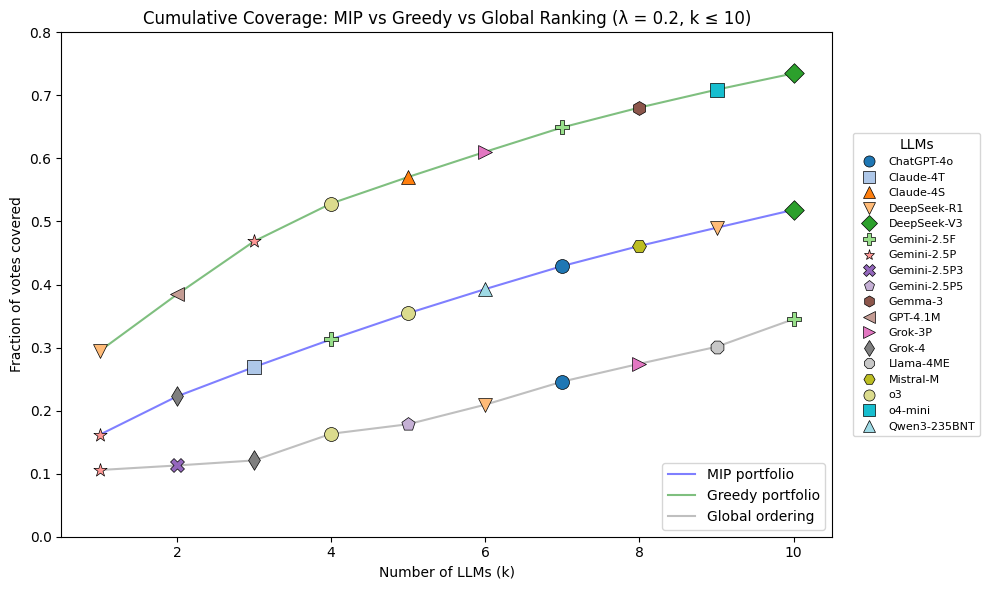}
        \caption{Cumulative vote coverage for $k = 10$ LLMs under the MIP portfolio, greedy portfolio, and global ordering at $\lambda = 0.20$.}
        \label{fig:coverage_comp_llms}
    \end{subfigure}
  
    \begin{subfigure}[t]{0.95\textwidth}
        \centering
    \includegraphics[width=0.9\textwidth]{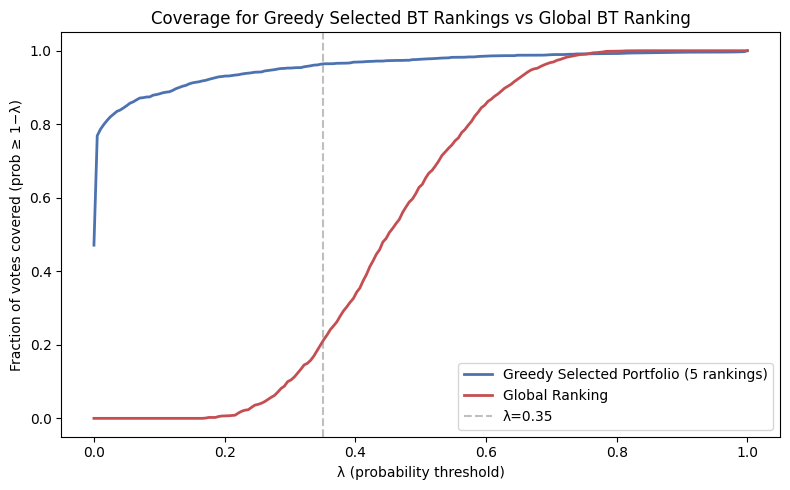}
        \caption{Vote coverage across $\lambda$ thresholds for the greedy-selected BT portfolio vs.\ the global BT ranking. Dashed line marks $\lambda = 0.35$.}
        \label{fig:greedy_vs_global_thresholds}
    \end{subfigure}
    \caption{Coverage by global ranking and by our portfolios.}\label{fig:arena_coverage_comparison}
\end{figure}

As Figure \ref{fig:arena_coverage_comparison} shows, the portfolio of $5$ rankings obtained by the greedy algorithm massively outperforms the global ranking in terms of coverage, for all margins $\lambda \in (0, 1)$, showing that the vast majority of all users can be put into one of $5$ `preference clusters'.

For $\nu = 0.95$, Tables \ref{tab:greedy-rankings} and \ref{tab:mip-rankings} respectively give the portfolio obtained by the greedy algorithm and \ref{ip: obj} respectively. We remark that MIP's portfolio outperforms or matches the greedy algorithm; e.g., for $\lambda = 0.50$, MIP obtains a portfolio of size $4$ while greedy algorithm obtains a portfolio of size $5$.

- \emph{Portfolios to LLMs.} Finally, we compute a portfolio of LLMs that together cover a large number of users using (1) the greedy algorithm's ranking portfolio, (2) MIP's ranking portfolio, and (3) using top-$k$ LLMs in the global ranking, for various values of $k$. The `error' of an LLM for a vote is computed based on a Bayesian approach, and depends on which of the original $389$ BT ranking cover the vote well and where the LLM ranks in those BT rankings (see Appendix \ref{app: bt-rankings-to-llm-scores} for details). Table~\ref{tab:poset-bt-rankings} describes the LLM portfolios obtained by the greedy algorithm and MIP's ranking portfolios.

\subsection{Extensions to COMPAS Dataset}\label{sec:compascase}

We briefly illustrate the same framework on the COMPAS dataset as an application to a classification setting, where the task is to predict recidivism within two years.

We describe our experimental setup for COMPAS. First, we build an ensemble of 61 classification models trained with different objective functions. One model minimizes standard binary cross-entropy (BCE) alone, while the remaining models incorporate a fairness regularization term as well: $\mathrm{BCE} + \mu \times \mathrm{EO}_{\mathrm{gap}}$ (equalized odds gap). In the latter objective, $\mu$ controls the weight of the fairness penalty and changes from 10 to 300, in increments of ten. The equalized odds gap ($\mathrm{EO}_{\mathrm{gap}}$) is defined as the sum of disparities in true positive rates (TPRs) and false positive rates (FPRs) across groups: $\mathrm{EO}_{\mathrm{gap}} := (\max_g TPR_g - \min_g TPR_g) + (\max_g FPR_g - \min_g FPR_g)$, computed separately over protected attribute groups. This results in two sets of fairness-regularized models: 30 models use sex-based groups and 30 models use race-based groups. Each model generates a predicted probability of 2-year recidivism over every individual $i$ in the dataset. Using these predictions, we compute $\lambda$-coverage at the individual level: a model $j$ covers an individual $i$ if the absolute difference between the predicted probability and the true binary outcome is at most $\lambda$ (i.e., $|h_j(i) - y_i| \le \lambda$); otherwise, the  individual is considered uncovered by that model.

\textbf{Results.} We show that a greedily-constructed portfolio of 4 models achieves 90\% coverage at $\lambda = 0.45$, with the selected models shown in Figure~\ref{fig:compas_greedy_coverage_by_lambda}. This suggests that even a relatively small and manageable set of models is sufficient to capture heterogeneity in the population. Moreover, models trained with different levels of fairness regularization appear to specialize in different subsets of individuals. At $\lambda = 0.4$, the greedily-constructed portfolio achieves 83.56\% coverage. Notably, the uncovered group comprises predominantly of younger individuals (under 45) with few prior offenses, and every individual is a repeat offender. These cases are missed by all models in the ensemble, including the global best model which only optimizes for BCE. This reflects a region of the feature space which is not well captured by the ensemble, indicating that such portfolios can be used to surface systematic blind spots in prediction coverage across both individuals and sub-populations, defined not only by protected attributes but also by behavioral characteristics.

\section{Conclusion}\label{sec: conclusion}
We show that global Bradley--Terry rankings on pairwise preference data from Arena are uninformative at the level of individual votes due to strong, structured heterogeneity across subpopulations. Some artefacts of this heterogeneity show up as a small spread of Elo scores, and model winning probabilities close to 0.53 (i.e, nearly random). One of our key insights is that {\it language} is the driving factor of LLM performance, conditional on language, the votes (hence opinions) are much more homogeneous. This insight may be of independent interest to language scholars, from the perspective of differences of LLMs performance based on language families (see Figure \ref{fig:heterogeneity}). Further, these trends may connect with literature on (1) observed differences in reasoning across languages (e.g., \cite{levinson2003space}, \cite{boroditsky2001does}), and (2) on the heterogeneity of opinions within each language family as their size (i.e., voters) grows (see Figure \ref{fig:global_error_lang_fam}, and e.g., \cite{kharkhurin2012multilingualism}). 

We propose $(\lambda,\nu)$-portfolios as an alternative to heterogeneous supervised ML settings, where a model ensemble can be sparsified so that each individual is well covered or their prediction error is within allowable $\lambda$ error. Small portfolios of BT rankings (and LLMs) achieve substantially higher vote coverage at comparable error thresholds. Similar gains hold in the COMPAS setting, suggesting that heterogeneous prediction problems are better served by a small set of specialized models. 

\textbf{Limitations.} Our analysis has some limitations, and leaves room for future work. In the Arena data, we do not observe voter identity, preventing separation of between-user heterogeneity from within-user noise. Further, although in applications like COMPAS it is easy to define subpopulations and fit models to them, in the Arena data, we assumed (and justified) that votes within a language-task subgroup can be meaningfully considered. However, despite our evidence of homogeneity within language and task-based groups, %it is unclear if these are \emph{optimal} predictors of user preferences.} 
we invite future analysis that searches for better vote clusters in this dataset, perhaps with more context from the users. Moreover, certain language families do not have votes from cross-pair comparisons across all LLMs, and this can prevent a meaningful total order\footnote{For e.g., consider the Italian $\times$ creative writing subpopulation. We observe votes for (Llama-4ME $>$ GPT-4.1M) and (o3 $>$ Nova-Exp), but none of the cross-pairs were compared. The BT model nonetheless assigns scores to all four and induces a total ordering Llama-4ME $>$ o3 $>$ Nova-Exp $>$ GPT-4.1M, despite no direct comparisons supporting the relative positions of models across the vote pairs.}.

\textbf{Broader impact.} Ranking systems shape decisions in science, technology, and society by distilling complex comparisons into simple leaderboards. While global rankings are appealing for their clarity and accessibility, our results show that they can obscure meaningful variation in preferences across populations, leading to potentially misleading conclusions.

We propose a simple alternative: small portfolios of models or rankings that better reflect diverse user needs while remaining interpretable. Beyond evaluation, our framework highlights populations that are poorly served and models that generalize broadly, offering a tool for auditing and more inclusive system design. These insights may be useful for policymakers and practitioners seeking to deploy AI systems in heterogeneous, real-world settings. 

\section{Acknowledgements} 
The authors would like to acknowledge the efforts of Akoua Orsot on  preliminary versions of the code on the COMPAS case study. 

\bibliographystyle{plainnat}
\bibliography{references, refs-zotero}

\newpage

\appendix

\section{Further details on Arena case-study}\label{app: arena-further-details}

Here, we supply further details for the Arena case-study in Section \ref{sec: arena}.

\subsection{Conversion table from Elo rating scale to prediction probability}

We include a table that maps the difference in Elo scores in a BT ranking and the BT ranking's confidence that the model with the higher score wins, up to $2$ decimal places. For example, for two models $a, b$ with scores $1050$ and $1450$ respectively (score difference of $400$), the BT ranking predicts that $a$ wins in a vote against $b$ with probability $\left(1 + \exp(\frac{1050 - 1450}{400})\right)^{-1} = 1/(1 + e^{-1}) \simeq 0.73$. Thus, the BT ranking predicts that $a$ wins against $b$ in roughly $3$ out of every $4$ votes. More generally, Eqn. \ref{eqn: elo-score} implies that an Elo rating difference of $\delta \ge 0$ implies that the model with higher score wins with probability
\[
    p(\delta) := \frac{1}{1 + \exp(-\delta)},
\]
with $p(0) = \frac{1}{2}$ (models with equal score are equally likely to win), and $\lim_{\delta \to \infty} p(\delta) = 1$.

\begin{table}[h]
    \centering
    \caption{Conversion between Elo score difference between two models and the predicted probability of the higher model winning in a vote.}
    \label{tab: elo-score-to-win-probability}
    \small
    \begin{tabular}{|l|ccccccccc|}
        \hline
        Elo score difference & 0 & 10 & 50 & 100 & 200 & 400 & 1000 & 1500 & 2000 \\
        \hline
        Win prob.\ (higher-ranked) & 0.50 & 0.51 & 0.53 & 0.56 & 0.62 & 0.73 & 0.92 & 0.98 & 0.99 \\
        \hline
    \end{tabular}
\end{table}

\subsection{LLM display name mapping}
\label{sec:llm-mapping}

Table~\ref{tab:llm-mapping} lists the full model identifiers and release dates for all 52 LLMs referenced in this work, alongside the abbreviated display names used in figures throughout the paper. Models are ordered by version/release date.

\begin{table}
    \centering
    \small
    \caption{Mapping of LLM display names to full model identifiers and version/release dates.\\}
    \label{tab:llm-mapping}
    \begin{tabular}{|l|l|l|}
        \toprule
        \textbf{Display name} & \textbf{Model identifier} & \textbf{Release date} \\
        \midrule
        GPT-4o-mini & \texttt{gpt-4o-mini-2024-07-18} & 2024-07-18 \\
        Claude-3.5H & \texttt{claude-3-5-haiku-20241022} & 2024-10-22 \\
        Claude-3.5S & \texttt{claude-3-5-sonnet-20241022} & 2024-10-22 \\
        GPT-4o & \texttt{gpt-4o-2024-11-20} & 2024-11-20 \\
        Nova-Pro & \texttt{amazon.nova-pro-v1:0} & 2024-12-03 \\
        Llama-3.3 & \texttt{llama-3.3-70b-instruct} & 2024-12-06 \\
        Gemini-2FT & \texttt{gemini-2.0-flash-thinking-exp-01-21} & 2025-01-21 \\
        Qwen-Max & \texttt{qwen-max-2025-01-25} & 2025-01-25 \\
        o3-mini & \texttt{o3-mini} & 2025-01-31 \\
        Gemini-2F & \texttt{gemini-2.0-flash-001} & 2025-02-05 \\
        Claude-3.7S & \texttt{claude-3-7-sonnet-20250219} & 2025-02-19 \\
        Claude-3.7ST & \texttt{claude-3-7-sonnet-20250219-thinking-32k} & 2025-02-19 \\
        Grok-3MB & \texttt{grok-3-mini-beta} & 2025-02-19 \\
        Grok-3MH & \texttt{grok-3-mini-high} & 2025-02-19 \\
        Grok-3P & \texttt{grok-3-preview-02-24} & 2025-02-24 \\
        QwQ-32B & \texttt{qwq-32b} & 2025-03-05 \\
        Gemma-3 & \texttt{gemma-3-27b-it} & 2025-03-10 \\
        Command-A & \texttt{command-a-03-2025} & 2025-03-13 \\
        Mistral-S-3.1 & \texttt{mistral-small-3.1-24b-instruct-2503} & 2025-03-17 \\
        DeepSeek-V3 & \texttt{deepseek-v3-0324} & 2025-03-24 \\
        Gemini-2.5P3 & \texttt{gemini-2.5-pro-preview-03-25} & 2025-03-25 \\
        ChatGPT-4o & \texttt{chatgpt-4o-latest-20250326} & 2025-03-26 \\
        Llama-4ME & \texttt{llama-4-maverick-03-26-experimental} & 2025-03-26 \\
        Llama-4MI & \texttt{llama-4-maverick-17b-128e-instruct} & 2025-04-05 \\
        Llama-4SI & \texttt{llama-4-scout-17b-16e-instruct} & 2025-04-05 \\
        GPT-4.1 & \texttt{gpt-4.1-2025-04-14} & 2025-04-14 \\
        GPT-4.1M & \texttt{gpt-4.1-mini-2025-04-14} & 2025-04-14 \\
        Hunyuan-Turbos & \texttt{hunyuan-turbos-20250416} & 2025-04-16 \\
        o3 & \texttt{o3-2025-04-16} & 2025-04-16 \\
        o4-mini & \texttt{o4-mini-2025-04-16} & 2025-04-16 \\
        Gemini-2.5-FP & \texttt{gemini-2.5-flash-preview-04-17} & 2025-04-17 \\
        Qwen3-235B & \texttt{qwen3-235b-a22b} & 2025-04-28 \\
        Qwen3-235BNT & \texttt{qwen3-235b-a22b-no-thinking} & 2025-04-28 \\
        Qwen3-30B & \texttt{qwen3-30b-a3b} & 2025-04-28 \\
        Gemini-2.5P5 & \texttt{gemini-2.5-pro-preview-05-06} & 2025-05-06 \\
        Mistral-M & \texttt{mistral-medium-2505} & 2025-05-07 \\
        Claude-4 & \texttt{claude-opus-4-20250514} & 2025-05-14 \\
        Claude-4T & \texttt{claude-opus-4-20250514-thinking-16k} & 2025-05-14 \\
        Claude-4S & \texttt{claude-sonnet-4-20250514} & 2025-05-14 \\
        Claude-4ST & \texttt{claude-sonnet-4-20250514-thinking-32k} & 2025-05-14 \\
        Nova-Exp & \texttt{amazon-nova-experimental-chat-05-14} & 2025-05-14 \\
        DeepSeek-R1 & \texttt{deepseek-r1-0528} & 2025-05-28 \\
        Magistral & \texttt{magistral-medium-2506} & 2025-06-10 \\
        Mistral-S & \texttt{mistral-small-2506} & 2025-06-10 \\
        Gemini-2.5F & \texttt{gemini-2.5-flash} & 2025-06-17 \\
        Gemini-2.5FLT & \texttt{gemini-2.5-flash-lite-preview-06-17-thinking} & 2025-06-17 \\
        Gemini-2.5P & \texttt{gemini-2.5-pro} & 2025-06-17 \\
        MiniMax & \texttt{minimax-m1} & 2025-06-17 \\
        Gemma-3n & \texttt{gemma-3n-e4b-it} & 2025-06-26\\
        Grok-4 & \texttt{grok-4-0709} & 2025-07-09 \\
        Kimi-K2 & \texttt{kimi-k2-0711-preview} & 2025-07-11 \\
        Qwen3-235BI & \texttt{qwen3-235b-a22b-instruct-2507} & 2025-07-21 \\
        \bottomrule
    \end{tabular}
\end{table}

\subsection{Language family mapping}
\label{sec:language-family}

Table~\ref{tab:lang-family} maps the 116 Arena languages to coarse language families used throughout the paper.

\begin{table}[t]
\centering
\small
\caption{Mapping of languages to language families. Groupings follow standard linguistic taxonomy (Glottolog) at a coarse level, e.g., Indo-Iranian combines the Indo-Aryan and Iranian branches, and isolate languages are grouped pragmatically.\\}
\label{tab:lang-family}
\setlength{\tabcolsep}{3pt}
\renewcommand{\arraystretch}{0.95}
\small
\begin{tabular}{@{}ll@{\hspace{12pt}}ll@{}}
\toprule
\textbf{Family} & \textbf{Languages} & \textbf{Family} & \textbf{Languages} \\
\midrule
Germanic & \parbox[t]{0.3\textwidth}{English, German, Dutch, Swedish, Norwegian, Norwegian Nynorsk, Danish, Icelandic, Afrikaans, Luxembourgish, Scots, Western Frisian} &
Afro-Asiatic & \parbox[t]{0.3\textwidth}{Arabic, Hausa, Somali, Maltese, Oromo, Afar} \\[6pt]
Romance & \parbox[t]{0.3\textwidth}{French, Spanish, Portuguese, Italian, Catalan, Romansh, Occitan, Galician, Corsican, Romanian, Haitian, Seselwa Creole French, Latin} &
Niger-Congo & \parbox[t]{0.3\textwidth}{Swahili, Zulu, Xhosa, Lingala, Wolof, Kinyarwanda, Ganda, Akan, Southern Sotho, Shona, Rundi, Sango} \\[6pt]
Slavic & \parbox[t]{0.3\textwidth}{Polish, Russian, Ukrainian, Czech, Slovak, Croatian, Bosnian, Serbian, Bulgarian, Belarusian, Macedonian, Slovenian} &
Austronesian & \parbox[t]{0.3\textwidth}{Malay, Tagalog, Cebuano, Malagasy, Waray, Tonga, Nauru, Indonesian} \\[6pt]
Indo-Iranian & \parbox[t]{0.3\textwidth}{Hindi, Urdu, Persian, Marathi, Gujarati, Bengali, Assamese, Sanskrit, Nepali, Sindhi, Kurdish, Tajik, Sinhala} &
Dravidian & \parbox[t]{0.3\textwidth}{Tamil, Malayalam, Kannada} \\[6pt]
Sino-Tibetan & \parbox[t]{0.3\textwidth}{Chinese, Tibetan, Burmese} &
Austroasiatic & \parbox[t]{0.3\textwidth}{Vietnamese, Khmer, Khasi} \\[6pt]
Turkic & \parbox[t]{0.3\textwidth}{Turkish, Uzbek, Azerbaijani, Kazakh, Kirghiz, Tatar, Uighur} &
Hellenic & \parbox[t]{0.3\textwidth}{Modern Greek} \\[6pt]
Uralic & \parbox[t]{0.3\textwidth}{Finnish, Estonian, Hungarian} &
Baltic & \parbox[t]{0.3\textwidth}{Latvian, Lithuanian} \\[6pt]
Celtic & \parbox[t]{0.3\textwidth}{Irish, Welsh, Manx} &
Kartvelian & \parbox[t]{0.3\textwidth}{Georgian} \\[6pt]
Koreanic & \parbox[t]{0.3\textwidth}{Korean} &
Albanian & \parbox[t]{0.3\textwidth}{Albanian} \\[6pt]
Japonic & \parbox[t]{0.3\textwidth}{Japanese} &
Armenian & \parbox[t]{0.3\textwidth}{Armenian} \\[6pt]
Tai-Kadai & \parbox[t]{0.3\textwidth}{Thai} &
Eskimo-Aleut & \parbox[t]{0.3\textwidth}{Kalaallisut} \\[6pt]
Isolate & \parbox[t]{0.3\textwidth}{Basque} &
Amerind & \parbox[t]{0.3\textwidth}{Quechua, Guarani} \\[6pt]
Constructed & \parbox[t]{0.3\textwidth}{Esperanto, Interlingue, Interlingua, Klingon, Volap\"uk} & & \\
\bottomrule
\end{tabular}
\end{table}

\subsection{Heterogeneity in language-specific rankings}
\label{sec:heterogeneity-in-rankings}

\begin{figure}[t]
    \centering
    \includegraphics[width=\linewidth]{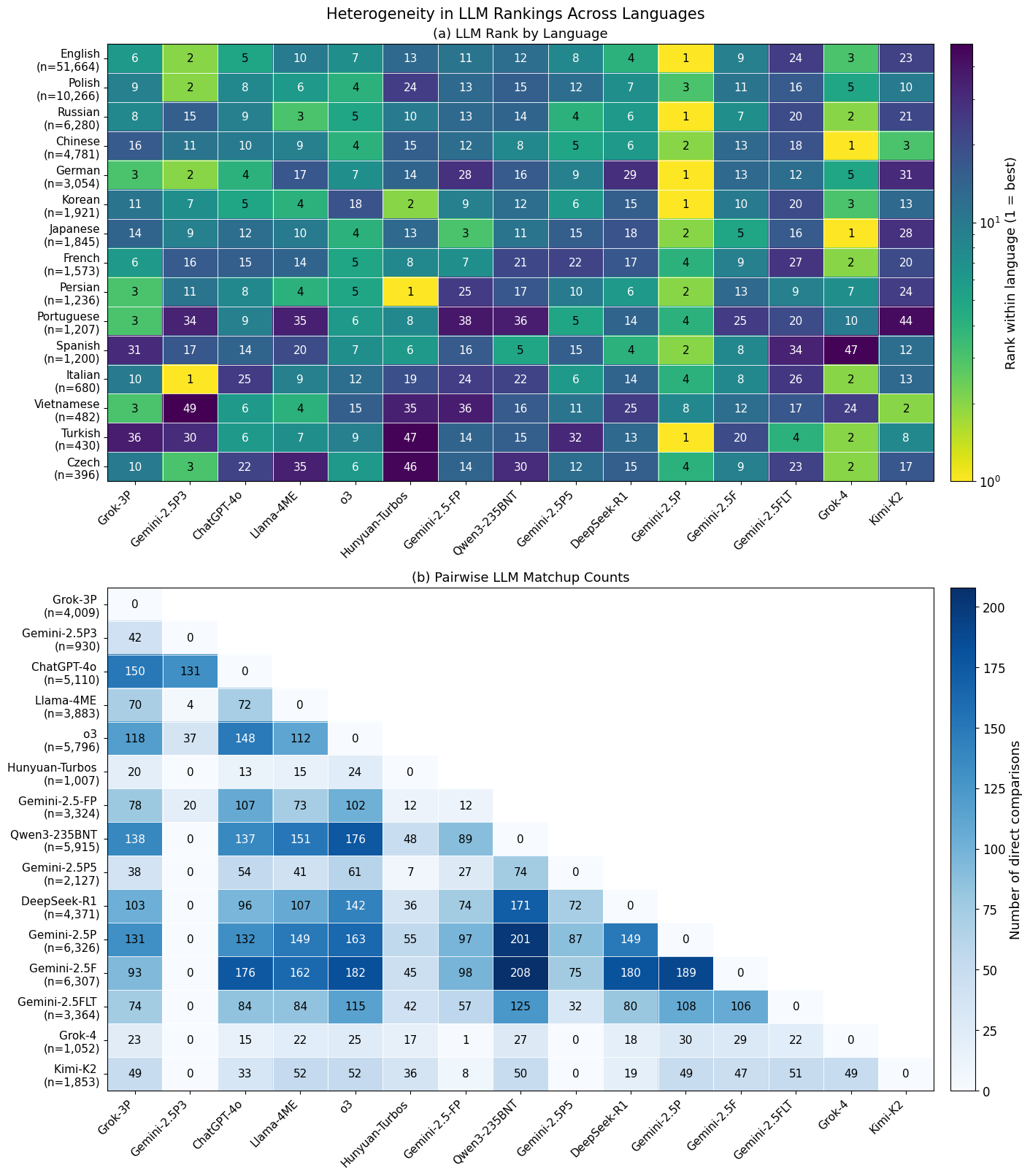}
    \caption{Heterogeneity in LLM rankings across languages. Rows: 15 languages with the most votes in Arena. Columns: top models by mean rank across these languages, ordered by release date. Each cell reports the model's within-language rank (1 = best). Demonstrates how the same model can rank in the top few for one language and well outside it for another.}
    \label{fig:heterogeneity}
\end{figure}

Figure \ref{fig:heterogeneity} shows ranks of top LLMs in the Arena dataset for the 15 most used languages in Arena.

\subsection{Portfolios of BT rankings}\label{sec:greedy-rankings}

\textbf{Greedy algorithm}. Table~\ref{tab:greedy-rankings} lists the greedy-selected BT ranking portfolios achieving $\nu = 0.95$ coverage at each $\lambda$ threshold, selected from 389 candidate rankings. Portfolio size $k$ is the number of rankings selected; coverage reports the fraction of 89{,}193 decisive votes $\lambda$-covered by at least one ranking in the portfolio.

\vspace{0.5em}
\begin{table}[t]
\begin{center}
%\captionsetup{type=table, font=small, skip=3pt}
\captionof{table}{The BT rankings selected by the greedy algorithm achieving $\nu = 0.95$ coverage at each $\lambda$ threshold.\\}
\label{tab:greedy-rankings}

% \scriptsize
\renewcommand{\arraystretch}{1.02}
\setlength{\tabcolsep}{3pt}
\begin{tabularx}{0.92\textwidth}{c c X}
\toprule
$\lambda$ & $k$ & Selected Rankings \\
\midrule
0.95 & 1 & global \\
0.90 & 1 & global \\
0.80 & 1 & English $\times$ code \\
0.70 & 1 & English $\times$ specificity \\
0.60 & 3 & English $\times$ specificity; Vietnamese $\times$ real\_world; Indonesian $\times$ complexity \\
0.50 & 5 & global; Portuguese $\times$ math; Portuguese $\times$ creative\_writing; Dutch $\times$ creativity; Persian $\times$ creative\_writing \\
0.45 & 4 & Persian $\times$ creative\_writing; Portuguese $\times$ creative\_writing; Romanian; Bengali $\times$ complexity \\
0.40 & 5 & Thai; Dutch $\times$ creativity; Japanese $\times$ math; Portuguese $\times$ creative\_writing; Romanian $\times$ domain\_knowledge \\
0.35 & 5 & Thai; Dutch $\times$ creativity; Japanese $\times$ math; Arabic $\times$ code; Romanian $\times$ domain\_knowledge \\
0.30 & 6 & Thai; Arabic $\times$ creative\_writing; Dutch $\times$ complexity; Arabic $\times$ code; Malay; Portuguese $\times$ math \\
0.25 & 6 & Hungarian $\times$ creativity; Arabic $\times$ creative\_writing; Arabic $\times$ code; Thai $\times$ domain\_knowledge; Dutch $\times$ complexity; Swedish $\times$ domain\_knowledge \\
0.20 & 6 & Arabic $\times$ creative\_writing; Thai; Arabic $\times$ code; Malay; Dutch $\times$ complexity; Swedish $\times$ domain\_knowledge \\
0.15 & 6 & Arabic $\times$ creative\_writing; Thai $\times$ domain\_knowledge; Afro-Asiatic $\times$ code; Hungarian $\times$ creativity; Malay; Hungarian $\times$ specificity \\
0.10 & 6 & Arabic $\times$ creative\_writing; Thai $\times$ domain\_knowledge; Arabic $\times$ code; Hungarian $\times$ creativity; Malay; Hungarian $\times$ specificity \\
0.05 & 7 & Arabic $\times$ creative\_writing; Thai $\times$ domain\_knowledge; Arabic $\times$ code; Hungarian $\times$ creativity; Malay; Latin; Swedish $\times$ domain\_knowledge \\
\bottomrule
\end{tabularx}
\end{center}
\end{table}
\vspace{0.5em}

% \section{MIP Selected Portfolio of BT Rankings}
% \label{sec:mip-rankings}

\textbf{MIP.} Table~\ref{tab:mip-rankings} lists the MIP-selected BT ranking portfolios achieving $\nu = 0.95$ coverage at each $\lambda$ threshold, selected from 389 candidate rankings. Portfolio size $k$ is the number of rankings selected; coverage reports the fraction of 89{,}193 decisive votes $\lambda$-covered by at least one ranking in the portfolio.

\begin{table}[t]
\begin{center}
\captionof{table}{The BT rankings selected by the MIP algorithm achieving $\nu = 0.95$ coverage at each $\lambda$ threshold.}
\label{tab:mip-rankings}
\renewcommand{\arraystretch}{1.02}
\setlength{\tabcolsep}{3pt}
\begin{tabularx}{0.92\textwidth}{c c X}
\toprule
$\lambda$ & $k$ & Selected Rankings \\
\midrule
0.95 & 1 & domain\_knowledge \\
0.90 & 1 & domain\_knowledge \\
0.80 & 1 & domain\_knowledge \\
0.70 & 1 & domain\_knowledge \\
0.60 & 3 & Japanese $\times$ technical\_accuracy, Indonesian $\times$ complexity, \\
     &   & Bengali $\times$ specificity \\
0.50 & 4 & Turkish $\times$ technical\_accuracy, Portuguese $\times$ creative\_writing, \\
     &   & Persian $\times$ creative\_writing, Romanian $\times$ domain\_knowledge \\
0.45 & 4 & Turkish $\times$ technical\_accuracy, Portuguese $\times$ creative\_writing, \\
     &   & Persian $\times$ creative\_writing, Romanian $\times$ domain\_knowledge \\
0.40 & 4 & Portuguese $\times$ creative\_writing, Persian $\times$ creative\_writing, \\
     &   & Romanian $\times$ domain\_knowledge, Other $\times$ specificity \\
0.35 & 5 & Thai, Portuguese $\times$ creative\_writing, Dutch $\times$ creativity, \\
     &   & Romanian $\times$ domain\_knowledge, Japonic $\times$ math \\
0.30 & 5 & Turkish $\times$ technical\_accuracy, Portuguese $\times$ creative\_writing, \\
     &   & Persian $\times$ creative\_writing, Romanian $\times$ domain\_knowledge, Other $\times$ specificity \\
0.25 & 5 & Danish $\times$ domain\_knowledge, Portuguese $\times$ creative\_writing, \\
     &   & Persian $\times$ creative\_writing, Romanian $\times$ domain\_knowledge, Other $\times$ specificity \\
0.20 & 6 & Malay, Portuguese $\times$ creative\_writing, Dutch $\times$ creativity, \\
     &   & Persian $\times$ creative\_writing, Indonesian $\times$ complexity, Romanian $\times$ domain\_knowledge \\
0.15 & 6 & Danish $\times$ domain\_knowledge, Dutch $\times$ technical\_accuracy, Persian $\times$ creative\_writing, \\
     &   & Arabic $\times$ code, Romanian $\times$ domain\_knowledge, Other $\times$ specificity \\
0.10 & 7 & Thai, Malay, Portuguese $\times$ creative\_writing, Japanese $\times$ math, \\
     &   & Czech $\times$ creative\_writing, Romanian $\times$ domain\_knowledge, Uralic $\times$ real\_world \\
0.05 & 7 & Thai, Malay, Serbian, Portuguese $\times$ creative\_writing, \\
     &   & Czech $\times$ creative\_writing, Romanian $\times$ domain\_knowledge, Uralic $\times$ real\_world \\
\bottomrule
\end{tabularx}
\end{center}
\end{table}

\subsection{BT rankings to LLM scores}\label{app: bt-rankings-to-llm-scores}

\newcommand{\score}{\mathrm{score}}
\newcommand{\E}{\mathbf{E}}

Having computed the set $R$ of $389$ BT rankings, we show how to compute the score of each LLM $\ell$ for any given vote $i = (a(i) \succ b(i))$. Recall that the LLM $\ell$ may not be one of $a_i$ or $b_i$.

For each ranking $r$, recall that the accuracy of the ranking for vote $i$ is defined as
\[
    p(i, r) := {\Pr}_r(a_i \succ b_i) = \frac{1}{1 + \exp(\theta^{(r)}_{b(i)} - \theta^{(r)}_{a(i)})}.
\]
This allows us to construct a probability distribution $\sigma(i)$ over the rankings, where $\sigma(i, r)$ indicates that vote $i$ is generated from ranking $r \in R$, or that the ranking $r$ represents the true preference of the corresponding user.
\[
    \sigma(i, r) := \frac{p(i, r)}{\sum_{s \in R} p(i, s)}.
\]
Then, we define the probability of an LLM $\ell$ satisfying the vote $i$ as:
\begin{align*}
    q(i, \ell) &:= {\Pr}_{\sigma(i)}(\ell \ \mathrm{is} \ \mathrm{at \ least \ as \ good \ as \ the \ winner \ } a_i) \\
    &= \sum_{r \in R} {\Pr}_{r}(\ell \ \mathrm{is} \ \mathrm{at \ least \ as \ good \ as \ the \ winner \ } a_i | i \mathrm{\ is \ generated \ from \ } r) \cdot \sigma(i, r) \\
    &= \begin{cases}
        \sum_{r \in R} \frac{\sigma(i, r)}{1 + \exp(\theta^{(r)}_{a(i)} - \theta^{(r)}_{\ell})} & \mathrm{if} \ \ell \neq a_i, \\
        1 & \mathrm{otherwise}.
    \end{cases}
\end{align*}

Finally, we define the `error' $\varepsilon(i, \ell)$ of LLM $\ell$ for vote $i$ as the probability of $\ell$ not beating $a_i$, i.e.,
\[
    \varepsilon(i, \ell) := 1 - q(i, \ell).
\]

\subsection{Pairwise win probability distributions}
\label{sec:win-prob-density}

\begin{figure}[H]
    \centering
    \includegraphics[width=0.92\textwidth]{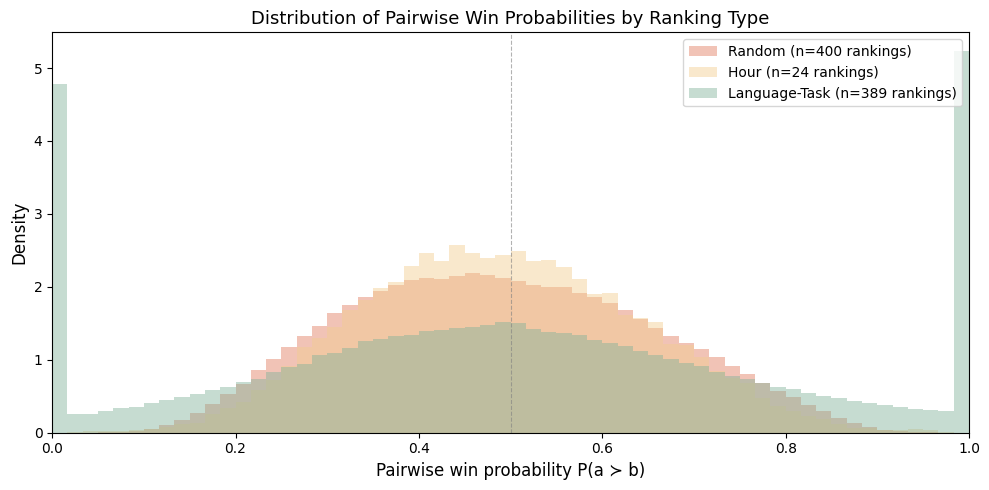}
    \caption{ Distribution of pairwise win probabilities $\Pr(a \succ b)$ by ranking type. Language-task rankings exhibit substantially more mass near 0 and 1 than random or hour-based baselines, indicating that stratified rankings induce sharper pairwise preferences.}
    \label{fig:win-prob-density}
\end{figure}

\subsection{Portfolios of LLMs}
\label{sec:poset-selected-bt-rankings}

Table~\ref{tab:poset-bt-rankings} compares the portfolios of LLMs induced by (1) greedy algorithm's portfolio of BT rankings, (2) MIP's portfolio of BT rankings, and (3) choosing the top-$k$ ranked models in the global BT ranking. The greedy and MIP portfolios are selected from the BT rankings at $\lambda = 0.05$ to achieve target coverage $\nu = 0.95$. The posets reported here are evaluated at $\lambda = 0.20$ to see how much cumulative coverage is achieved with $k = 10$ LLMs. The global baseline corresponds to the global BT ordering.

\begin{table}[t]
\centering
\caption{Model orderings induced by the greedy-selected, global, and MIP-selected BT ranking portfolios.}
\label{tab:poset-bt-rankings}
\setlength{\tabcolsep}{3pt}
\renewcommand{\arraystretch}{1}

\begin{minipage}[t]{0.45\textwidth}
\centering
\textbf{Greedy portfolio}
\begin{tabularx}{\textwidth}{c X c}
\toprule
Rank & LLM & Coverage \\
\midrule
1 & \seqsplit{deepseek-r1-0528} & 0.295 \\
2 & \seqsplit{gpt-4.1-mini-2025-04-14} & 0.384 \\
3 & \seqsplit{gemini-2.5-pro} & 0.468 \\
4 & \seqsplit{o3-2025-04-16} & 0.528 \\
5 & \seqsplit{claude-sonnet-4-20250514} & 0.571 \\
6 & \seqsplit{grok-3-preview-02-24} & 0.611 \\
7 & \seqsplit{gemini-2.5-flash} & 0.649 \\
8 & \seqsplit{gemma-3-27b-it} & 0.681 \\
9 & \seqsplit{o4-mini-2025-04-16} & 0.709 \\
10 & \seqsplit{deepseek-v3-0324} & 0.735 \\
\bottomrule
\end{tabularx}
\end{minipage}
\hfill
\begin{minipage}[t]{0.45\textwidth}
\centering
\textbf{Global ordering}
\begin{tabularx}{\textwidth}{c X c}
\toprule
Rank & LLM & Coverage \\
\midrule
1 & \seqsplit{gemini-2.5-pro} & 0.106 \\
2 & \seqsplit{gemini-2.5-pro-preview-03-25} & 0.113 \\
3 & \seqsplit{grok-4-0709} & 0.121 \\
4 & \seqsplit{o3-2025-04-16} & 0.163 \\
5 & \seqsplit{gemini-2.5-pro-preview-05-06} & 0.178 \\
6 & \seqsplit{deepseek-r1-0528} & 0.209 \\
7 & \seqsplit{chatgpt-4o-latest-20250326} & 0.246 \\
8 & \seqsplit{grok-3-preview-02-24} & 0.274 \\
9 & \seqsplit{llama-4-maverick-03-26-experimental} & 0.301 \\
10 & \seqsplit{gemini-2.5-flash} & 0.345 \\
\bottomrule
\end{tabularx}
\end{minipage}
\hfill
\vspace{0.2cm}
\noindent 
\begin{minipage}[t]{0.45\textwidth}
\centering
\textbf{MIP portfolio}
\begin{tabularx}{\textwidth}{c X c}
\toprule
Rank & LLM & Coverage \\
\midrule
1 & \seqsplit{gemini-2.5-pro} & 0.162 \\
2 & \seqsplit{grok-4-0709} & 0.223 \\
3 & \seqsplit{claude-opus-4-20250514-thinking-16k} & 0.269 \\
4 & \seqsplit{gemini-2.5-flash} & 0.313 \\
5 & \seqsplit{o3-2025-04-16} & 0.354 \\
6 & \seqsplit{qwen3-235b-a22b-no-thinking} & 0.393 \\
7 & \seqsplit{chatgpt-4o-latest-20250326} & 0.429 \\
8 & \seqsplit{mistral-medium-2505} & 0.461 \\
9 & \seqsplit{deepseek-r1-0528} & 0.490 \\
10 & \seqsplit{deepseek-v3-0324} & 0.519 \\
\bottomrule
\end{tabularx}
\end{minipage}
\end{table}

\subsection{Coverage of language-specific votes by BT rankings}
\label{sec:global-bt-coverage-language}

\begin{figure}[t]
\centering
\includegraphics[width=0.9\linewidth]{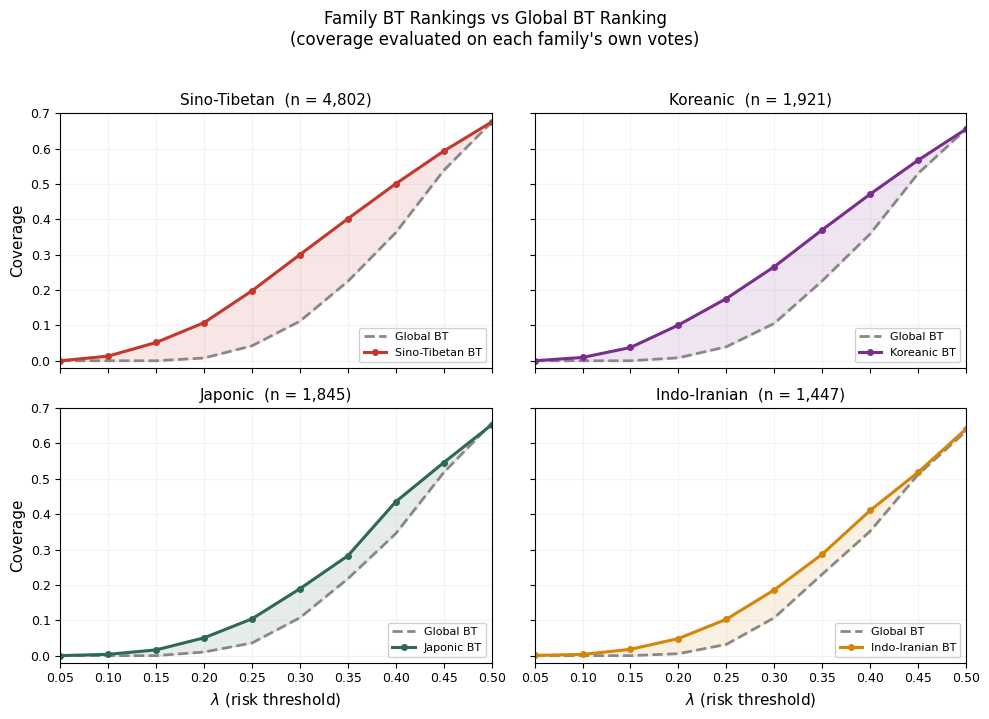}
\par\vspace{1cm}
\begin{tabular}{r@{\hspace{0.5em}}l@{\hspace{0.5em}}r}
            \toprule
            \textbf{Rank} & \textbf{Model} & \textbf{BT Score (Elo-scaled)} \\
            \midrule
            1  & Gemini-2.5P    & 1144 \\
            2  & Gemini-2.5P3   & 1114 \\
            3  & Grok-4         & 1093 \\
            4  & o3             & 1093 \\
            5  & Gemini-2.5P5   & 1091 \\
            6  & DeepSeek-R1    & 1088 \\
            7  & ChatGPT-4o     & 1088 \\
            8  & Grok-3P        & 1080 \\
            9  & Llama-4ME      & 1074 \\
            10 & Gemini-2.5F    & 1072 \\
            11 & Gemini-2.5-FP  & 1052 \\
            12 & Qwen3-235BNT   & 1044 \\
            13 & Hunyuan-Turbos & 1044 \\
            14 & Kimi-K2        & 1039 \\
            15 & Qwen3-235B     & 1036 \\
            \midrule
            48 & Mistral-S-3.1  & 892 \\
            49 & Nova-Pro       & 890 \\
            50 & GPT-4o-mini    & 876 \\
            51 & Claude-3.5H    & 871 \\
            52 & Magistral      & 867 \\
            \bottomrule
        \end{tabular}

\caption{Language Family BT rankings achieve better coverage than a global BT ranking. The top panel compares coverage under language-family-specific BT rankings against the global BT ranking, while the bottom table reports the top 15 models under the global BT ranking along with their BT scores.}
\label{fig:family_comp}
\end{figure}

\begin{figure}
    \centering
    \includegraphics[width=\textwidth]{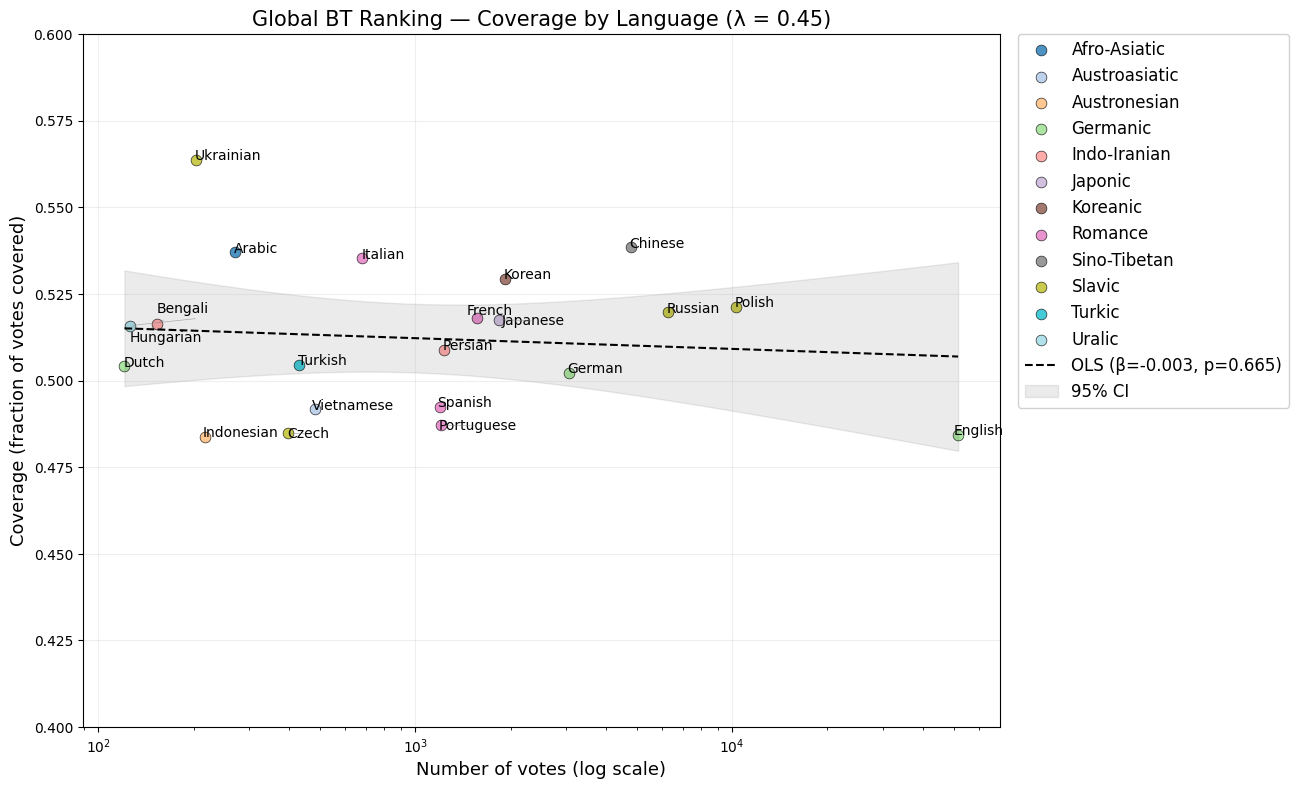}
    \caption{Coverage of the global BT ranking by language at $\lambda = 0.45$. Each point corresponds to a language that has at least 100 votes, colored by language family. The x-axis shows the number of votes on a log scale, and the y-axis shows the fraction of votes covered by the global ranking. The variation across languages indicates that a single global ranking does not represent all language subpopulations uniformly. The dashed line shows an ordinary least squares fit, with the shaded region indicating the 95\% confidence interval.
    The weak slope (-0.003) suggests that variation in global-ranking coverage is not simply explained by the number of votes available for each language.}
    \label{fig:global_error_lang_fam}
\end{figure}

Figure~\ref{fig:global_error_lang_fam} reports the coverage of the global BT ranking separately for each language at $\lambda = 0.45$. Each point represents one language, with the x-axis showing the number of votes on a log scale and the y-axis showing the fraction of votes covered.

% \section{Portfolio Error Distributions}
% \label{sec:portfolio-errors}

\subsection{Portfolio error distributions}
\label{sec:portfolio-errors}

\begin{figure}
    \centering
    \includegraphics[width=\textwidth]{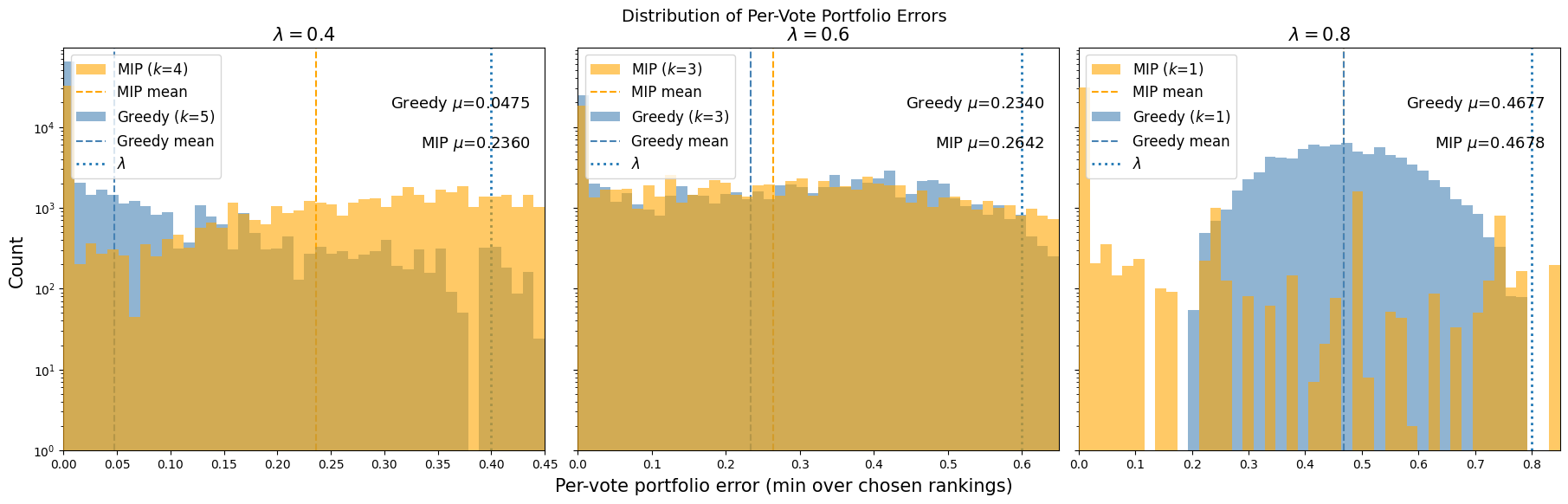}
    \caption{Per-vote portfolio error distributions for MIP and greedy BT ranking portfolios at $\lambda \in \{0.4, 0.6, 0.8\}$. Portfolio error for each vote is the minimum error across selected rankings. Dashed lines: mean error; dotted lines: $\lambda$ threshold.}
    \label{fig:portfolio-errors}
\end{figure}

Figure~\ref{fig:portfolio-errors} compares per-vote portfolio errors for the greedy and MIP portfolios. At tighter thresholds (e.g., $\lambda = 0.4$), the greedy portfolio concentrates more error mass near zero. Its first few picks target high-coverage language strata, yielding low error on the votes they cover, but leave a heavier tail of poorly covered votes. This results in a higher overall mean error than the MIP portfolio. As $\lambda$ increases, this gap narrows and at $\lambda = 0.8$, the two methods achieve nearly identical mean errors. Greedy excels at covering easy votes cheaply, while MIP distributes coverage more evenly, so hybrid selection strategies may be a promising direction for future work.

\section{Further details on COMPAS case-study}\label{app: compas-further-details}

Here, we supply further details for the COMPAS case-study in Section \ref{sec:compascase}.

\subsection{Model portfolio-sizes by $\lambda$ values for different population coverage targets}
\label{sec:compas-coverage}

\begin{figure}
    \centering\includegraphics[width=0.8\textwidth]{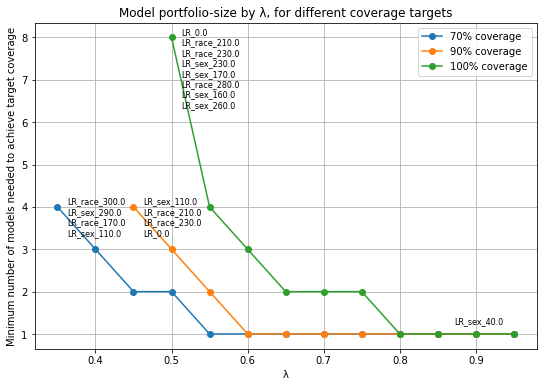}
    \caption{Minimum number of models needed to achieve target population coverage across $\lambda$ thresholds.The models in the greedily-constructed portfolios are listed for the largest portfolio-size for a given coverage target and for when the portfolio converges to a single model.}
    \label{fig:compas_greedy_coverage_by_lambda}
\end{figure}

Figure~\ref{fig:compas_greedy_coverage_by_lambda} shows the minimum number of models required to achieve a target coverage level over the full population for different values of $\lambda$ where coverage is attainable. As expected, higher coverage targets require larger model portfolios and are achievable at looser thresholds (higher $\lambda$ values). For example, to achieve 100\% coverage, we need 8 models at $\lambda$ = 0.5, whereas 70\% coverage only requires 2 models at the same $\lambda$ threshold. Similarly, portfolios converge to a single model at lower $\lambda$ thresholds for lower coverage targets.

\subsection{Analysis of uncovered population}
\label{sec:compas-uncovered}

Table \ref{tab:compas_lambda_0.4} shows the portfolio constructed at $\lambda = 0.4$. Here, the model portfolio achieves 83.56\% coverage, with the remaining individuals assigned to “No model.” We explicitly retain this uncovered set to analyze the characteristics of individuals not captured by any model in the ensemble. We find that this group is predominantly younger individuals with few prior offenses, across all race and sex groups, and that every individual in this subset is a repeat offender (as shown in Table \ref{tab:compas_lambda_0.4_uncovered_F} for the female sub-population). This highlights that the portfolio approach can also be used as a tool for surfacing blind spots in predictive systems, perhaps offering policymakers insight into where models may systematically under-perform and, as a result, where greater attention in oversight might be required.

\begin{table}[!t]
\centering
\caption{Assignments to model portfolio for the COMPAS data constructed at $\lambda = 0.4$, by sub-population. Data point that were not covered by any regularized model are depicted in the last row.\\}
\label{tab:compas_lambda_0.4}
\begin{tabular}{l c c c c c c}
\toprule
\textbf{Model} &
\textbf{F-AA} &
\textbf{M-AA} &
\textbf{F-C} &
\textbf{M-C} &
\textbf{F-O} &
\textbf{M-O} \\
\midrule
Global & 0 & 0 & 0 & 11 & 4 & 0 \\
Race(100) & 1 & 0 & 0 & 0 & 6 & 0 \\
Race(140) & 0 & 0 & 0 & 1 & 0 & 0 \\
Race(150) & 0 & 0 & 0 & 1 & 0 & 0 \\
Race(180) & 0 & 0 & 0 & 1 & 1 & 0 \\
Race(210) & 8 & 8 & 3 & 32 & 24 & 5 \\
Race(230) & 2 & 7 & 2 & 6 & 9 & 4 \\
Race(300) & 17 & 15 & 4 & 78 & 49 & 18 \\
Sex(10) & 0 & 2 & 0 & 0 & 0 & 0 \\
Sex(110) & 69 & 41 & 17 & 277 & 180 & 76 \\
Sex(140) & 1 & 0 & 0 & 2 & 2 & 0 \\
Sex(180) & 0 & 0 & 0 & 0 & 2 & 0 \\
Sex(240) & 0 & 0 & 0 & 0 & 2 & 0 \\
Sex(290) & 5 & 3 & 0 & 22 & 5 & 9 \\
No model & 13 & 10 & 5 & 94 & 57 & 24 \\
\bottomrule
\end{tabular}

\vspace{0.3em}

\parbox{\linewidth}{\small
Notes: F-AA = Female African-American, M-AA = Male African-American, F-C = Female Caucasian, M-C = Male Caucasian, F-O = Female Other, M-O = Male Other.

Global model is the best fit linear regression model, with no regularization. The Sex($\mu$) and Race($\mu$) depict fairness-regularized models with penalty parameter $\mu$ set as indicated.

}
\end{table}

\begin{table}[ht]
%\centering
\caption{Features of the uncovered female sub-population, using the constructed portfolio at $\lambda$ = 0.4\\}
\label{tab:compas_lambda_0.4_uncovered_F}

\small
\setlength{\tabcolsep}{5pt}
\renewcommand{\arraystretch}{1.5}

\begin{tabularx}{\textwidth}{c c c c c c c c c c}
\toprule
\textbf{Age} & 
\textbf{\makecell{Age\\Category}} & 
\textbf{Sex} & 
\textbf{Race} &
\textbf{\makecell{Priors\\Count}} & 
\textbf{\makecell{Days}} & 
\textbf{\makecell{Length\\of Stay}} &
\textbf{\makecell{Charge\\Type}} & 
\textbf{\makecell{2-Year\\Recid}} &  \\
\midrule

24 & Less than 25 & Female & African-American & 0 & -1.0 & 0 & F & 1 &  \\
21 & Less than 25 & Female & African-American & 0 & -1.0 & 1 & F & 1 &  \\
23 & Less than 25 & Female & African-American & 0 & -1.0 & 3 & F & 1 &  \\
23 & Less than 25 & Female & African-American & 0 & -1.0 & 2 & F & 1 &  \\
21 & Less than 25 & Female & African-American & 0 & 0.0 & 0 & F & 1 &  \\
24 & Less than 25 & Female & Caucasian & 0 & -1.0 & 0 & M & 1 &  \\
30 & 25 - 45 & Female & Other & 0 & -1.0 & 1 & M & 1 &  \\
32 & 25 - 45 & Female & Other & 0 & -1.0 & 7 & M & 1 &  \\
29 & 25 - 45 & Female & Other & 0 & -1.0 & 1 & F & 1 &  \\
29 & 25 - 45 & Female & Other & 0 & -2.0 & 1 & M & 1 &  \\
21 & Less than 25 & Female & Other & 0 & -1.0 & 0 & M & 1 &  \\
20 & Less than 25 & Female & African-American & 1 & -1.0 & 1 & F & 1 &  \\
21 & Less than 25 & Female & African-American & 1 & -1.0 & 0 & F & 1 &  \\
26 & 25 - 45 & Female & Caucasian & 1 & -1.0 & 0 & F & 1 &  \\
34 & 25 - 45 & Female & Caucasian & 1 & -1.0 & 12 & F & 1 &  \\
33 & 25 - 45 & Female & Caucasian & 1 & -5.0 & 4 & M & 1 &  \\
31 & 25 - 45 & Female & African-American & 2 & -1.0 & 0 & F & 1 &  \\
23 & Less than 25 & Female & African-American & 2 & -2.0 & 1 & F & 1 &  \\
24 & Less than 25 & Female & African-American & 2 & -1.0 & 1 & F & 1 & \\
39 & 25 - 45 & Female & Caucasian & 2 & -1.0 & 0 & F & 1 & \\
31 & 25 - 45 & Female & Caucasian & 2 & -1.0 & 1 & F & 1 &  \\
29 & 25 - 45 & Female & Caucasian & 2 & -1.0 & 8 & F & 1 &  \\
27 & 25 - 45 & Female & African-American & 3 & -1.0 & 0 & M & 1 &  \\
38 & 25 - 45 & Female & African-American & 3 & 0.0 & 0 & F & 1 &  \\
44 & 25 - 45 & Female & Caucasian & 3 & 0.0 & 11 & F & 1 & \\
33 & 25 - 45 & Female & Caucasian & 4 & -1.0 & 11 & M & 1 &  \\
53 & Greater than 45 & Female & Caucasian & 4 & -1.0 & 0 & F & 1 &  \\
35 & 25 - 45 & Female & African-American & 5 & -1.0 & 7 & F & 1 &\\

\bottomrule
\end{tabularx}

\vspace{0.2cm}

\small{
Features related to juvenile criminal history counts are not shown as they are primarily zero across the entire sub-population.\\
Days denotes the number of days between screening and arrest.\\
Charge-type M indicates misdemeanor.\\
Charge-type F indicates felony.\\
No model was able to cover these defendants within reasonable error.
}

\end{table}

\subsection{FPR comparison between single best model and greedily-constructed portfolios}
\label{sec:compas-fpr}
In Table \ref{tab:portfolio_fpr_results}, we compare model portfolios constructed under different $\lambda$ thresholds for full population coverage against a global best model that optimizes only BCE. As expected, portfolios constructed at lower $\lambda$ thresholds have larger sizes, reflecting a stricter notion of coverage that requires a more diverse set of models to account for variation across individuals. Under $\lambda$ = 0.4, the constructed portfolio achieves zero overall and sub-population FPRs, suggesting that a wider, more diverse set of models improves the ability to capture differences across individuals, leading to more precise predictions.

As $\lambda$ increases, smaller portfolios can achieve the targeted coverage (100\%) but the resulting portfolios have non-zero FPRs across the full population and within most sub-populations. This indicates that the selected model sets no longer capture all of the individual nuances. Despite this, we find that the portfolio approach consistently yields lower FPRs than the global best model for the overall population. Within the sub-populations, the  FPRs from portfolios are also generally lower than those of the global model for all sub-populations and at all  $\lambda$ values, with the exception of where $\lambda$ = 0.8. In this case, where a single model can be used to cover the entire population, the FPR increases slightly for the Male-Caucasian sub-population compared to the global best model but remains lower for the overall and all other sub-population FPRs.

\begin{table}[!t]
\caption{False Positive Rate (FPR) metrics across sub-populations for different model portfolios}
\label{tab:portfolio_fpr_results}
\setlength{\tabcolsep}{6pt}
\renewcommand{\arraystretch}{1.3}
\small
\begin{tabularx}{\textwidth}{>{\raggedright\arraybackslash}p{2.5cm}cp{1.2cm}p{1cm}p{0.8cm}p{0.8cm}p{0.8cm}p{0.8cm}p{0.8cm}p{0.8cm}}
\toprule
\textbf{Model Portfolio} &
\textbf{$\lambda$} &
\textbf{Target Coverage} &
\textbf{Overall FPR} &
\textbf{F-AA} &
\textbf{M-AA} &
\textbf{F-C} &
\textbf{M-C} &
\textbf{F-O} &
\textbf{M-O} \\
\midrule

\{Global\} & N/A & N/A & 0.216 & 0.103 & 0.388 & 0.043 & 0.178 & 0.012 & 0.107 \\

\{Sex(150)\} 
& 0.8 & 100\%  & 0.103 & 0.014 & 0.056 & 0.167 & 0.198 & 0.048 & 0.037 \\

\{Race(30), Sex(10)\}
%\{$LR_{\text{race}-30.0}$, $LR_{\text{sex}-10.0}$\} 
& 0.7 & 100\%  & 0.094 & 0.014 & 0.130 & 0.093 & 0.090 & 0.000 & 0.098 \\

\{Race(30, 40), Sex(140)\}
%$LR_{\text{race}-30.0}$, $LR_{\text{race}-40.0}$, $LR_{\text{sex}-140.0}$\} 
& 0.6 & 100\%  & 0.074 & 0.014 & 0.091 & 0.093 & 0.080 & 0.000 & 0.073 \\

\{Global, $\text{Race}(210, 230, 280)$, Sex$(160, 170, 230, 260)$\}%_{\text{race}-210.0}$, $LR_{\text{race}-230.0}$, $LR_{\text{race}-280.0}$, $LR_{\text{sex}-160.0}$, $LR_{\text{sex}-170.0}$, $LR_{\text{sex}-230.0}$, $LR_{\text{sex}-260.0}$\}  
& 0.5 & 100\%  & 0.000 & 0.000 & 0.000 & 0.000 & 0.000 & 0.000 & 0.000 \\

%\{$LR_{\text{sex}-150.0}$\} 
\bottomrule
\end{tabularx}

\vspace{0.2cm}

\footnotesize{
Reported FPR values are rounded to three decimal places.

Global model is the best fit linear regression model, with no regularization. The Sex($\mu$) and Race($\mu$) depict fairness-regularized models with penalty parameter $\mu$ set as indicated. 

F-AA = Female African-American, 
M-AA = Male African-American, 
F-C = Female Caucasian, \\
M-C = Male Caucasian, 
F-O = Female Other, 
M-O = Male Other.
}

\end{table}

\section{Implementation details for \ref{ip: obj}}\label{sec: mip-implementation-details}

Since \ref{ip: obj} is an integer program, it may not be tractable to solve to optimality in a reasonable time. Instead, we solve the linear relaxation optimally, and obtain fractional variables $u_i, t_j \in [0, 1]$ for data points $i$ and models $j$. We use the following \emph{continuous threshold} rounding scheme to round these to an integer solution and obtain a portfolio:
\begin{itemize}
    \item initialize portfolio $P \gets \emptyset$ of models
    \item while fewer than $\nu n$ points have been $\lambda$-covered by models in $P$, choose the model not in $P$ that has the highest value of $t_j$, and add it to $P$
\end{itemize}

When each $t_j$ is integral, it is easy to check that this returns the corresponding portfolio $P := \{j: t_j = 1\}$. Further, we also run a second phase of the MIP to obtain smaller overall errors: after the rounding, suppose the MIP returns a portfolio of size $|P| = k$, showing that a $(\lambda, \nu)$-portfolio of size $k$ exists. Then, we write another mixed-integer program to find the $(\lambda, \nu)$-portfolio of fixed size $k$ that minimizes the overall mean-squared error across data points, where each data point is assigned to the model in the portfolio that has the smallest error on the data point. 

\section{Data sources}\label{sec: data-sources}

The Arena dataset data was provides by Arena on HuggingFace (\url{https://huggingface.co/datasets/lmarena-ai/arena-human-preference-140k}) under the CC BY 4.0 license. COMPAS data was obtained from ProPublica (\url{https://github.com/propublica/compas-analysis}) under the CC BY 4.0 license.

\end{document}